\documentclass{article}

\usepackage[final,nonatbib]{nips_2016}

\usepackage[utf8]{inputenc} 
\usepackage[T1]{fontenc}    
\usepackage{url}            
\usepackage{booktabs}       
\usepackage{amsfonts}       
\usepackage{nicefrac}       
\usepackage{microtype}      
\usepackage{caption}

\usepackage{algorithm}
\usepackage{algorithmic}
\usepackage{amsmath}
\usepackage{amsthm}
\usepackage{amssymb}
\usepackage{graphicx}
\usepackage{xspace}
\usepackage{tabularx}
\usepackage{multicol}
\usepackage{multirow}
\usepackage{url}
\usepackage{wrapfig}
\usepackage{comment}
\usepackage{listings}
\usepackage[utf8]{inputenc}
\usepackage{fancyvrb}
\usepackage{booktabs}
\usepackage{color,soul}
\usepackage[table]{xcolor}
\usepackage[normalem]{ulem}

\newcommand{\bmx}[0]{\begin{bmatrix}}
\newcommand{\emx}[0]{\end{bmatrix}}

\newcommand{\matr}[1]{\mathbf{#1}}

\newcommand{\mI}{\matr{I}}

\newcommand{\RR}[0]{\mathbb{R}}

\newcommand{\safe}[0]{\ensuremath{\text{safe}}}
\newcommand{\supervised}[0]{\ensuremath{\text{supervised}}}

\newcommand{\dagg}[0]{\ensuremath{\text{DAgger}}}

\DeclareMathOperator*{\argmax}{\arg \max}
\DeclareMathOperator*{\argmin}{\arg \min}

\title{Query-Efficient Imitation Learning for  \\ 
End-to-End Autonomous Driving}

\author{
  Jiakai Zhang \\
  Department of Computer Science\\
  New York University\\
  \texttt{zhjk@nyu.edu} \\
  \And
  Kyunghyun Cho \\
  Department of Computer Science\\
  Center for Data Science\\
  New York University \\
  \texttt{kyunghyun.cho@nyu.edu} \\
}

\begin{document}

\maketitle

\begin{abstract}

    One way to approach end-to-end autonomous driving is to learn a policy
    function that maps from a sensory input, such as an image frame from a
    front-facing camera, to a driving action, by imitating an expert driver, or
    a reference policy.  This can be done by supervised learning, where a policy
    function is tuned to minimize the difference between the predicted and
    ground-truth actions. A policy function trained in this way however is known
    to suffer from unexpected behaviours due to the mismatch between the states
    reachable by the reference policy and trained policy functions.  More
    advanced algorithms for imitation learning, such as DAgger, addresses this
    issue by iteratively collecting training examples from both reference and
    trained policies. These algorithms often requires a large number of queries
    to a reference policy, which is undesirable as the reference policy is often
    expensive. In this paper, we propose an extension of the DAgger, called
    SafeDAgger, that is query-efficient and more suitable for end-to-end
    autonomous driving. We evaluate the proposed SafeDAgger in a car racing
    simulator and show that it indeed requires less queries to a reference
    policy. We observe a significant speed up in convergence, which we
    conjecture to be due to the effect of automated curriculum learning.

\end{abstract}

\section{Introduction}

We define end-to-end autonomous driving as driving by a single, self-contained
system that maps from a sensory input, such as an image frame from a
front-facing camera, to actions necessary for driving, such as the angle of
steering wheel and braking. In this approach, the autonomous driving system is
often {\it learned} from data rather than manually designed, mainly due to sheer
complexity of manually developing a such system. 

This end-to-end approach to autonomous driving dates back to late 80's. ALVINN
by Pomerleau~\cite{pomerleau1989alvinn} was a neural network with a single
hidden layer that takes as input an image frame from a front-facing camera and
a response map from a range finder sensor and returns a quantized steering wheel
angle. The ALVINN was trained using a set of training tuples (image, sensor map,
steering angle) collected from simulation. A similar approach was taken later in
2005 to train, this time, a convolutional neural network to drive an off-road
mobile robot~\cite{muller2005off}. More recently,
Bojarski~et~al.~\cite{bojarski2016nvidia} used a similar, but deeper,
convolutional neural network for lane following based solely on a front-facing
camera. In all these cases, a deep neural network has been found to be
surprisingly effective at learning a complex mapping from a raw image to
control.

A major learning paradigm behind all these previous attempts has been supervised
learning. A human driver or a rule-based AI driver in a simulator, to which we
refer as a reference policy drives a car equipped with a front-facing camera and
other types of sensors while collecting image-action pairs. These collected
pairs are used as training examples to train a neural network controller, called
a primary policy. It is however well known that a purely supervised learning
based approach to imitation learning (where a learner tries to imitate a human
driver) is suboptimal (see, e.g., \cite{daume2009search,ross2010reduction} and
references therein.)

We therefore investigate a more advanced approach to imitation learning for
training a neural network controller for autonomous driving. More specifically,
we focus on DAgger~\cite{ross2010reduction}  which works in a setting where the
reward is given only implicitly. DAgger improves upon supervised learning by
letting a primary policy collect training examples while running a reference
policy simultaneously. This dramatically improves the performance of a neural
network based primary policy.  We however notice that DAgger needs to constantly
query a reference policy, which is expensive especially when a reference policy
may be a human driver. 

In this paper, we propose a query-efficient extension of the DAgger, called {\it
SafeDAgger}. We first introduce a safety policy that learns to predict the error
made by a primary policy {\it without} querying a reference policy. This safety
policy is incorporated into the DAgger's iterations in order to select only a
small subset of training examples that are collected by a primary policy. This
subset selection significantly reduces the number of queries to a reference
policy. 

We empirically evaluate the proposed SafeDAgger using TORCS~\cite{torcs}, a
racing car simulator, which has been used for vision-based autonomous driving
research in recent years~\cite{koutnik2013evolving,chen2015deepdriving}. In this
paper, our goal is to learn a primary policy that can drive a car indefinitely
without any crash or going out of a road.  The experiments show that the
SafeDAgger requires much less queries to a reference policy than the original
DAgger does and achieves a superior performance in terms of the average number
of laps without crash and the amount of damage. We conjecture that this is due
to the effect of automated curriculum learning created by the subset selection
based on the safety policy.


\section{Imitation Learning for Autonomous Driving}

In this section, we describe imitation learning in the context of learning an
automatic policy for driving a car. 

\subsection{State Transition and Reward}

A surrounding environment, or a world, is defined as a set of states $S$. Each
state is accompanied by a set of possible actions $A(S)$. Any given state $s\in
S$ transitions to another state $s'\in S$ when an action $a \in A(S)$ is
performed, according to a state transition function $\delta: S \times A(S) \to S$.
This transition function may be either deterministic or stochastic. 

For each sequence of state-action pairs, there is an associated (accumulated) reward $r$: 
\begin{align*}
    r(\Omega=((s_0, a_0), (s_1, a_1), (s_2, a_2), \ldots)),
\end{align*}
where 
$
    s_t = \delta(s_{t-1}, a_{t-1}).
    $

A reward may be implicit in the sense that the reward comes as a form of a
binary value with 0 corresponding to any unsuccessful run (e.g., crashing into
another car so that the car breaks down,) while any successful run (e.g.,
driving indefinitely without crashing) does not receive the reward.  This is the
case in which we are interested in this paper. In learning to drive, the reward
is simply defined as follows:
\begin{align*}
    r(\Omega) = \left\{
        \begin{array}{l l}
            1, & \text{ if there was no crash}, \\
            0, & \text{ otherwise}
        \end{array}
        \right.
\end{align*}
This reward is implicit, because it is observed only when there is a failure,
and no reward is observed with an optimal policy (which never crashes and drives
indefinitely.)

\subsection{Policies}

A policy is a function that maps from a state observation $\phi(s)$ to one $a$
of the actions available $A(s)$ at the state $s$. An underlying state $s$
describes the surrounding environment perfectly, while a policy often has only a
limited access to the state via its observation $\phi(s)$. In the context of
end-to-end autonomous driving, $s$ summarizes all necessary information about
the road (e.g., \# of lanes, existence of other cars or pedestrians, etc.,)
while $\phi(s)$ is, for instance, an image frame taken by a front-facing camera.

We have two separate policies. First, a {\it primary policy} $\pi$ is a policy
that learns to drive a car. This policy does not observe a full, underlying
state $s$ but only has access to the state observation $\phi(s)$, which is in
this paper a pixel-level image frame from a front-facing camera. The primary
policy is implemented as a function parametrized by a set of parameters
$\theta$. 

The second one is a {\it reference policy} $\pi^*$. This policy may or may not
be optimal, but is assumed to be a good policy which we want the primary policy
to {\it imitate}. In the context of autonomous driving, a reference policy can
be a human driver. We use a rule-based controller, which has access to a true,
underlying state in a driving simulator, as a reference policy in this paper.

\paragraph{Cost of a Policy}
Unlike previous works on imitation learning (see, e.g.,
\cite{daume2009search,ross2010reduction,chang2015learning}), we introduce a
concept of cost to a policy. The cost of querying a policy given a state for an
appropriate action varies significantly based on how the policy is implemented.
For instance, it is expensive to query a reference policy, if it is a human
driver. On the other hand, it is much cheaper to query a primary policy which is
often implemented as a classifier. Therefore, in this paper, we analyze an
imitation learning algorithm in terms of {\it how many queries it makes to a
reference policy}. 

\subsection{Driving}

A car is driven by querying a policy for an action with a state observation
$\phi(s)$ at each time step. The policy, in this paper, observes an image frame
from a front-facing camera and returns both the angle of a steering wheel ($u
\in \left[-1, 1\right]$) and a binary indicator for braking ($b \in \left\{ 0,
1\right\}$). We call this strategy of relying on a single fixed policy a {\it
naive strategy}.

\paragraph{Reachable States}
With a set of initial state $S_0^\pi \subset S$, each policy $\pi$ defines a
subset of the reachable states $S^\pi$. That is, 
$
    S^\pi = \cup_{t=1}^{\infty} S^\pi_t,
$
where 
$
    S^\pi_t = \left\{ 
        s | s = \delta(s', \pi(\phi(s')))~~\forall s' \in S^\pi_{t-1}
    \right\}.
$
In other words, a car driven by a policy $\pi$ will only visit the states in
$S^\pi$. 

We use $S^*$ to be a reachable set by the reference policy. In the case of
learning to drive, this reference set is intuitively smaller than that by any
other reasonable, non-reference policy. This happens, as the reference policy
avoids any state that is likely to lead to a low reward which corresponds to
crashing into other cars and road blocks or driving out of the road.

\subsection{Supervised Learning}

Imitation learning aims at finding a primary policy $\pi$ that {\it imitates} a
reference policy $\pi^*$. The most obvious approach to doing so is supervised
learning. In supervised learning, a car is first driven by a reference policy
while collecting the state observations $\phi(s)$ of the visited states,
resulting in 
$
    D = \left\{ \phi(s)_1, \phi(s)_2, \ldots, \phi(s)_N \right\}.
$
Based on this dataset, we define a loss function as
\begin{align}
    \label{eq:sup_cost1}
    l_\supervised(\pi, \pi^*, D) = \frac{1}{N} \sum_{n=1}^N \| \pi(\phi(s)_n) -
    \pi^*(\phi(s)_n) \|^2.
\end{align}
Then, a desired primary policy is
$
    \hat{\pi} = \argmin_{\pi} l_\supervised(\pi, \pi^*, D).
$

A major issue of this supervised learning approach to imitation learning stems
from the imperfection of the primary policy $\hat{\pi}$ even after training.
This imperfection likely leads the primary policy to a state $s$ which is not
included in the reachable set $S^*$ of the reference policy, i.e., $s \notin
S^*$. As this state cannot have been included in the training set $D \subseteq
S^*$, the behaviour of the primary policy becomes unpredictable. The
imperfection arises from many possible factors, including sub-optimal loss
minimization, biased primary policy, stochastic state transition and partial
observability.

\subsection{DAgger: beyond Supervised Learning}
\label{sec:dagger}

A major characteristics of the supervised learning approach described above is
that it is only the reference policy $\pi^*$ that generates training examples.
This has a direct consequence that the training set is almost a subset of the
reference reachable set $S^*$.  The issue with supervised learning can however
be addressed by imitation learning or
learning-to-search~\cite{daume2009search,ross2010reduction}. 

In the framework of imitation learning, the primary policy, which is currently
being estimated, is also used in addition to the reference policy when
generating training examples. The overall training set used to tune the primary
policy then consists of both the states reachable by the reference policy as
well as the intermediate primary policies. This makes it possible for the
primary policy to correct its path toward a good state, when it visits a state
unreachable by the reference policy, i.e., $s \in S^\pi \backslash S^*$.

DAgger is one such imitation learning algorithm proposed in
\cite{ross2010reduction}. This algorithm finetunes a primary policy trained
initially with the supervised learning approach described earlier. Let $D_0$ and
$\pi_0$ be the supervised training set (generated by a reference policy) and the
initial primary policy trained in a supervised manner. Then, DAgger iteratively
performs the following steps. At each iteration $i$, first, additional training
examples are generated by a mixture of the reference $\pi^*$ and primary
$\pi_{i-1}$ policies (i.e., 
\begin{align}
    \label{eq:col_policy}
    \beta_i \pi^* + (1 - \beta_i) \pi_{i-1}
\end{align}
) and
combined with all the previous training sets:
$
    D_i = D_{i-1} \cup \left\{ \phi(s)_1^i, \ldots, \phi(s)_N^i \right\}.
$
The primary policy is then finetuned, or trained from scratch, by minimizing
$l_\supervised(\theta, D_i)$ (see Eq.~\eqref{eq:sup_cost1}.) This iteration
continues until the supervised cost on a validation set stops improving.

DAgger does not rely on the availability of explicit reward. This makes it
suitable for the purpose in this paper, where the goal is to build an end-to-end
autonomous driving model that drives on a road indefinitely. However, it is
certainly possible to incorporate an explicit reward with other imitation
learning algorithms, such as SEARN~\cite{daume2009search},
AggreVaTe~\cite{ross2014reinforcement} and LOLS~\cite{chang2015learning}.
Although we focus on DAgger in this paper, our proposal later on applies
generally to any learning-to-search type of imitation learning algorithms.

\paragraph{Cost of DAgger}

At each iteration, DAgger queries the reference policy for each and every
collected state. In other words, the cost of DAgger $C^\dagg_i$ at the $i$-th
iteration is equivalent to the number of training examples collected, i.e,
$C^\dagg_i = \left| D_i \right|$. In all, the cost of DAgger for learning a
primary policy is $C^\dagg = \sum_{i=1}^M \left| D_i \right|$, excluding the initial
supervised learning stage. 

This high cost of DAgger comes with a more practical issue, when a reference
policy is a human operator, or in our case a human driver. First, as noted in
\cite{ross2013learning}, a human operator cannot drive well without actual
feedback, which is the case of DAgger as the primary policy drives most of the
time. This leads to suboptimal labelling of the collected training examples.
Furthermore, this constant operation easily exhausts a human operator, making it
difficult to scale the algorithm toward more iterations. 

\section{SafeDAgger: Query-Efficient Imitation Learning with a Safety Policy}

We propose an extension of DAgger that minimizes the number of queries to a
reference policy both during training and testing. In this section, we describe
this extension, called {\it SafeDAgger}, in detail.

\subsection{Safety Policy}
\label{sec:safety_policy}

Unlike previous approaches to imitation learning, often as
learning-to-search~\cite{daume2009search,ross2010reduction,chang2015learning},
we introduce an additional policy $\pi_\safe$, to which we refer as a {\it
safety policy}. This policy takes as input both the partial observation of a
state $\phi(s)$ and a primary policy $\pi$ and returns a binary label indicating
whether the primary policy $\pi$ is likely to deviate from a reference policy
$\pi^*$ {\it without} querying it. 

We define the deviation of a primary policy $\pi$ from a reference policy
$\pi^*$ as
\begin{align*}
    \epsilon(\pi, \pi^*, \phi(s)) = \left\| \pi(\phi(s))- \pi^*(\phi(s)) \right\|^2.
\end{align*}
Note that the choice of error metric can be flexibly chosen depending on a
target task. For instance, in this paper, we simply use the $L_2$ distance
between a reference steering angle and a predicted steering angle, ignoring the
brake indicator.

Then, with this defined deviation, the optimal safety policy $\pi_\safe^*$ is
defined as 
\begin{align}
    \label{eq:safe_policy}
    \pi^*_\safe(\pi, \phi(s)) = \left\{
        \begin{array}{l l}
            0, & \text{ if } \epsilon(\pi, \pi^*, \phi(s)) > \tau \\
            1, & \text{ otherwise} 
        \end{array}
        \right.,
\end{align}
where $\tau$ is a predefined threshold. The safety policy decides whether the
choice made by the policy $\pi$ at the current state can be trusted with respect
to the reference policy. We emphasize again that this determination is done
without querying the reference policy.


\paragraph{Learning}
A safety policy is {\it not} given, meaning that it needs to be estimated during
learning. A safety policy $\pi_\safe$ can be learned by collecting another set of
training examples:\footnote{
    It is certainly possible to simply set aside a subset of the original
    training set for this purpose.
}
$
    D' = \left\{ \phi(s)_1', \phi(s)_2', \ldots, \phi(s)_N' \right\}.
$
We define and minimize a binary
cross-entropy loss:
\begin{align}
    \label{eq:safe_cost}
    l_\safe(\pi_\safe, \pi, \pi^*, D') = - \frac{1}{N} \sum_{n=1}^N  &
    \pi_\safe^*(\phi(s)_n') \log \pi_\safe(\phi(s)_n', \pi)
    +  \\ &
    (1 - \pi_\safe^*(\phi(s)_n')) \log (1- \pi_\safe(\phi(s)_n', \pi)),
    \nonumber
\end{align}
where we model the safety policy as returning a Bernoulli distribution over
$\left\{ 0, 1\right\}$. 

\paragraph{Driving: Safe Strategy}

Unlike the naive strategy, which is a default go-to strategy in most cases of
reinforcement learning or imitation learning, we can design a {\it safe
strategy} by utilizing the proposed safety policy $\pi_\safe$. In this strategy,
at each point in time, the safety policy determines whether it is safe to let
the primary policy drive. If so (i.e., $\pi_\safe(\pi, \phi(s)) = 1$,) we use
the action returned by the primary policy (i.e., $\pi(\phi(s))$.) If not (i.e.,
$\pi_\safe(\pi, \phi(s)) = 0$,) we let the reference policy drive instead (i.e.,
$\pi^*(\phi(s))$.) 

Assuming the availability of a good safety policy, this strategy avoids any
dangerous situation arisen by an imperfect primary policy, that may lead to a
low reward (e.g., break-down by a crash.) In the context of learning to drive,
this safe strategy can be thought of as letting a human driver take over the
control based on an automated decision.\footnote{
    Such intervention has been done manually by a human driver~\cite{pomerleau1992progress}.
}
Note that this driving strategy is
applicable regardless of a learning algorithm used to train a primary policy.

\paragraph{Discussion}
The proposed use of safety policy has a potential to address this issue up to a
certain point. First, since a separate training set is used to train the safety
policy, it is more robust to unseen states than the primary policy. Second and
more importantly, the safety policy finds and exploits a simpler decision
boundary between safe and unsafe states instead of trying to learn a complex
mapping from a state observation to a control variables.  For instance, in
learning to drive, the safety policy may simply learn to distinguish between a
crowded road and an empty road and determine that it is safer to let the primary
policy drive in an empty road.


\paragraph{Relationship to a Value Function}

A value function $V^\pi(s)$ in reinforcement learning computes the reward a
given policy $\pi$ can achieve in the future starting from a given state $s$
\cite{sutton1998reinforcement}. This description already reveals a clear
connection between the safety policy and the value function. The safety policy
$\pi_\safe(\pi, s)$ determines whether a given policy $\pi$ is likely to {\it
fail} if it operates at a given state $s$, in terms of the deviation from a
reference policy. By assuming that a reward is only given at the very end of a
policy run and that the reward is $1$ if the current policy acts exactly like
the reference policy and otherwise $0$, the safety policy precisely returns the
value of the current state. 

A natural question that follows is whether the safety policy can drive a car on
its own. This perspective on the safety policy as a value function suggests a
way to using the safety policy directly to drive a car. At a given state $s$,
the best action $\hat{a}$ can be selected to be
$
    \argmax_{a \in A(s)} \pi_\safe(\pi, \delta(s, a)).
$
This is however not possible in the current formulation, as the transition
function $\delta$ is unknown. 
We may extend the definition of the
proposed safety policy so that it considers a state-action pair $(s,a)$ instead
of a state alone and predicts the safety in the next time step, which makes it
closer to a $Q$ function. 

\begin{algorithm}[t]
    \caption{SafeDAgger {\scriptsize {\color{blue} Blue} fonts are used to highlight the differences
    from the vanilla DAgger.}}
    \label{alg:safedagger}
    \begin{algorithmic}[1]
        \STATE Collect $D_0$ using a reference policy $\pi^*$
        \STATE {\color{blue} Collect $D_\safe$ using a reference policy $\pi^*$}
        \STATE $\pi_0 = \argmin_{\pi} l_\supervised(\pi, \pi^*, D_0)$ 
        \STATE {\color{blue} $\pi_{\safe,0} = \argmin_{\pi_\safe} l_\safe(\pi_\safe, \pi_0, \pi^*, D_\safe \cup D_0)$}
        \FOR{$i=1$ \TO $M$}
        \STATE Collect $D'$ {\color{blue} using the {\bf safety strategy} using
            $\pi_{i-1}$ and $\pi_{\safe,i-1}$}
            \STATE {\color{blue} {\bf Subset Selection}: $D' \leftarrow \left\{
                    \phi(s) \in D' |
            \pi_{\safe,i-1}(\pi_{i-1}, \phi(s)) = 0 \right\}$}
            \STATE $D_i = D_{i-1} \cup D'$
            \STATE $\pi_i = \argmin_{\pi} l_\supervised(\pi, \pi^*, D_i)$ 
            \STATE {\color{blue} $\pi_{\safe,i} = \argmin_{\pi_\safe} l_\safe(\pi_\safe, \pi_i, \pi^*, D_\safe \cup D_i)$}
        \ENDFOR
        \RETURN $\pi_M$ {\color{blue} and $\pi_{\safe,M}$}
    \end{algorithmic}
    
\end{algorithm}

\subsection{SafeDAgger: Safety Policy in the Loop}

We describe here the proposed SafeDAgger which aims at reducing the number of
queries to a reference policy during iterations. At the core of SafeDAgger lies
the safety policy introduced earlier in this section. The SafeDAgger is
presented in Alg.~\ref{alg:safedagger}. There are two major modifications to the
original DAgger from Sec.~\ref{sec:dagger}.

First, we use the safe strategy, instead of the naive strategy, to collect
training examples (line 6 in Alg.~\ref{alg:safedagger}). This allows an agent to simply give up when it is not safe to
drive itself and hand over the control to the reference policy, thereby
collecting training examples with a much further horizon without crashing. This
would have been impossible with the original DAgger unless the manually forced
take-over measure was implemented \cite{ross2013learning}. 


Second, the subset selection (line 7 in Alg.~\ref{alg:safedagger}) drastically
reduces the number of queries to a reference policy. Only a small subset of
states where the safety policy returned $0$ need to be labelled with reference
actions. This is contrary to the original DAgger, where all the collected states
had to be queried against a reference policy. 

Furthermore, this subset selection allows the subsequent supervised learning to
focus more on difficult cases, which almost always correspond to the states that
are problematic (i.e., $S \backslash S^*$.) This reduces the total amount of
training examples without losing important training examples, thereby making
this algorithm data-efficient. 

Once the primary policy is updated with $D_i$ which is a union of the initial
training set $D_0$ and all the hard examples collected so far, we update the
safety policy. This step ensures that the safety policy correctly identifies
which states are difficult/dangerous for the latest primary policy. This has an
effect of automated curriculum learning~\cite{bengio2009curriculum} with a mix
strategy~\cite{zaremba2014learning}, where the safety policy selects training
examples of appropriate difficulty at each iteration.

Despite these differences, the proposed SafeDAgger inherits much of the
theoretical guarantees from the DAgger. This is achieved by gradually increasing
the threshold $\tau$ of the safety policy (Eq.~\eqref{eq:safe_policy}). If $\tau
> \epsilon(\pi, \phi(s))$ for all $s \in S$, the SafeDAgger reduces to the
original DAgger with $\beta_i$ (from Eq.~\eqref{eq:col_policy}) set to $0$.  We
however observe later empirically that this is not necessary, and that training
with the proposed SafeDAgger with a fixed $\tau$ automatically and gradually
reduces the portion of the reference policy during data collection over
iterations. 

\paragraph{Adaptation to Other Imitation Learning Algorithms} The proposed use
of a safety policy is easily adaptable to other more recent cost-sensitive
algorithms. In AggreVaTe~\cite{ross2014reinforcement}, for instance, the
roll-out by a reference policy may be executed not from a uniform-randomly
selected time point, but from the time step when the safety policy returns $0$.
A similar adaptation can be done with LOLS~\cite{chang2015learning}. We do not
consider these algorithms in this paper and leave them as future work.

\section{Experimental Setting}

\subsection{Simulation Environment}

We use TORCS~\cite{torcs}, a racing car simulator, for empirical evaluation in
this paper. We chose TORCS based on the following reasons. First, it has been
used widely and successfully as a platform for research on autonomous
racing~\cite{loiacono2008wcci}, although most of the previous work, except for
\cite{koutnik2013evolving,chen2015deepdriving},  are not comparable as they use
a radar instead of a camera for observing the state.  Second, TORCS is a
light-weight simulator that can be run on an off-the-shelf workstation. Third,
as TORCS is an open-source software, it is easy to interface it with another
software which is Torch in our case.\footnote{
    We will release a patch to TORCS that allows seamless integration between
    TORCS and Torch.
}

\paragraph{Tracks}

To simulate a highway driving with multiple lanes, 
we modify the original TORCS road surface textures by adding various lane
configurations such as the number of lanes, the type of lanes. We use ten tracks
in total for our experiments. We split those ten tracks into two disjoint sets:
seven training tracks and three test tracks. All training examples as well as 
validation examples are collected from the
training tracks only, and a trained primary policy is tested on the test tracks.
See Fig.~1 for the visualizations of the tracks and Appendix~\ref{sec:dataset} for the types of
information collected as examples.

\paragraph{Reference Policy $\pi^*$}

We implement our own reference policy which has access to an underlying state
configuration. The state includes the position, heading direction, speed, and
distances to others cars. The reference policy either follows the current lane
(accelerating up to the speed limit), changes the lane if there is a slower car
in the front and a lane to the left or right is available, or brakes.  

\subsection{Data Collection}

We use a car in TORCS driven by a policy to collect data. For each training track, 
we add 40 cars driven by the reference policy to simulate traffic.  
We run up to three iterations in addition to the initial supervised learning
stage. In the case of SafeDAgger, we collect 30k, 30k and 10k of training
examples (after the subset selection in line 6 of Alg.~\ref{alg:safedagger}.) In
the case of the original DAgger, 
we collect up to 390k data each iteration and uniform-randomly select 30k, 30k
and 10k of training examples. 



\subsection{Policy Networks}

\paragraph{Primary Policy $\pi_\theta$}

\begin{wrapfigure}{R}{0.32\textwidth}
    \centering
    \includegraphics[width=\linewidth]{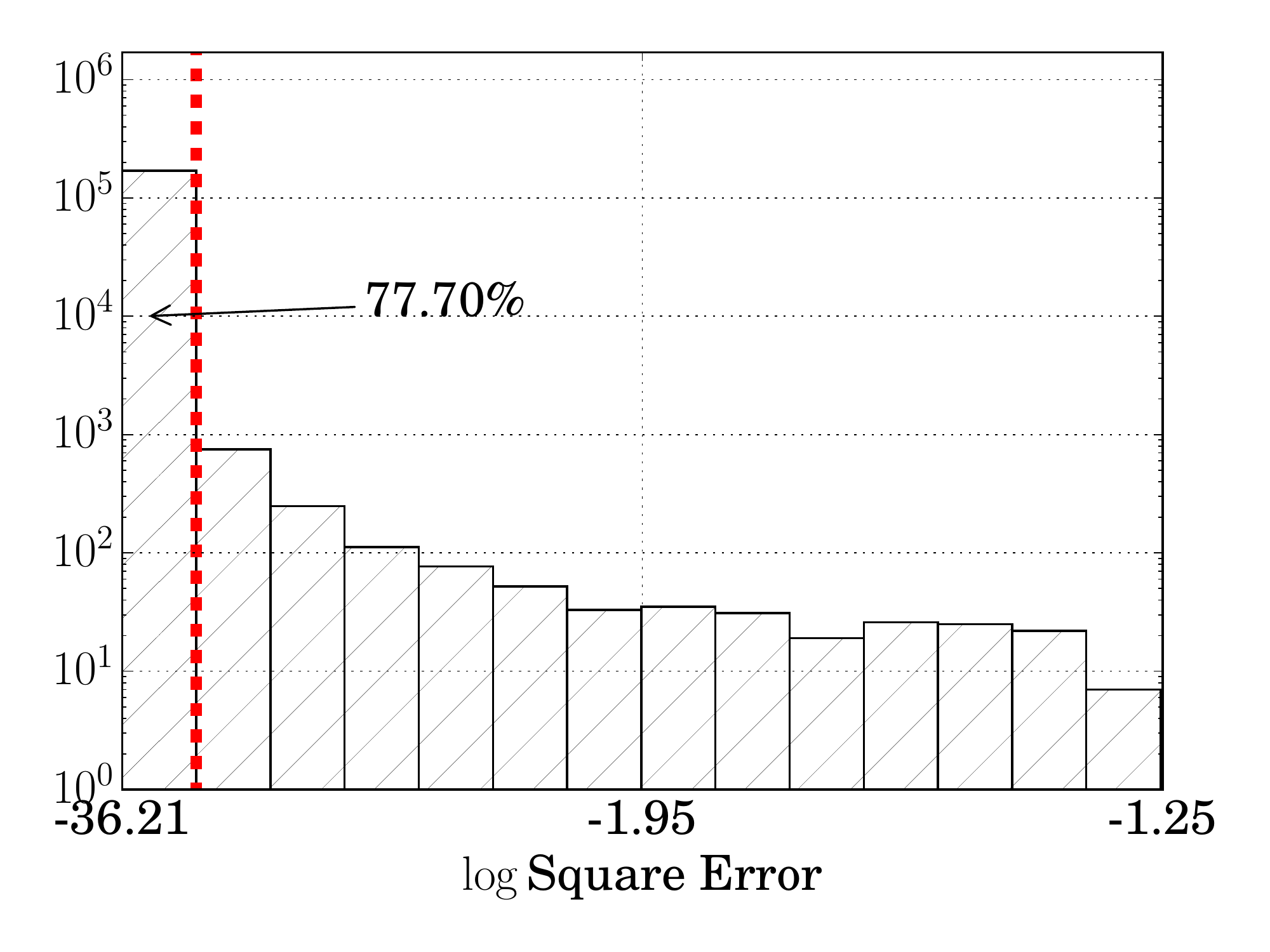}

    \caption{The histogram of the $\log$ square errors of steering angle after
        {\it supervised} learning only.
        The dashed line is located at $\tau=0.0025$. $77.70\%$ of the training
    examples are considered safe.}
    \label{fig:safetyThr}
    \vspace{-8mm}
\end{wrapfigure}

We use a deep convolutional network that has five convolutional layers
followed by a set of fully-connected layers. This convolutional network takes
as input the pixel-level image taken from a front-facing camera. It predicts the
angle of steering wheel ($\left[-1, 1\right]$) and whether to brake ($\left\{ 0,
1\right\}$). Furthermore, the network predicts as an auxiliary task the car's
affordances, including the existence of a lane to the left or right of the car
and the existence of another car to the left, right or in front of the car. 
We have found this multi-task approach to easily outperform a single-task network,
confirming the promise of multi-task learning from \cite{caruana1997multitask}.

\paragraph{Safety Policy $\pi_\safe$}

We use a feedforward network to implement a safety policy.  The activation of
the primary policy network's last hidden convolutional layer is fed through two
fully-connected layers followed by a softmax layer with two categories
corresponding to $0$ and $1$. We choose $\tau=0.0025$ as our safety policy
threshold so that approximately 20\% of initial training examples are considered
unsafe, as shown in Fig.~\ref{fig:safetyThr}. See Fig.~\ref{fig:frames} in the Appendix
for some examples of which frames were determined safe or unsafe.

For more details, see Appendix~\ref{sec:policy} in the Appendix.

\subsection{Evaluation}

\paragraph{Training and Driving Strategies}

We mainly compare three training strategies; (1)Supervised Learning, (2) DAgger 
(with $\beta_i = \mI_{i=0}$)
and (3) SafeDAgger. For each training strategy, we evaluate trained policies 
with both of the driving strategies; (1) naive strategy and (2) safe strategy.

\paragraph{Evaluation Metrics}

We evaluate each combination by letting it drive on the three test tracks up to
three laps. All these runs are repeated in two conditions; without traffic and
with traffic, while recording three metrics. The first metric is {\it the number of completed laps} without
going outside a track, averaged over the three tracks. When a car drives out of
the track, we immediately halt. Second, we look at a damage accumulated while
driving.  Damage happens each time the car bumps into another car. Instead of a
raw, accumulated damage level, we report the {\it damage per lap}. Lastly, we
report the {\it mean squared error of steering angle}, computed
while the primary policy drives.

\section{Results and Analysis}

\begin{figure}
    \centering

    \begin{minipage}{0.32\textwidth}
        \centering
        \includegraphics[width=\columnwidth]{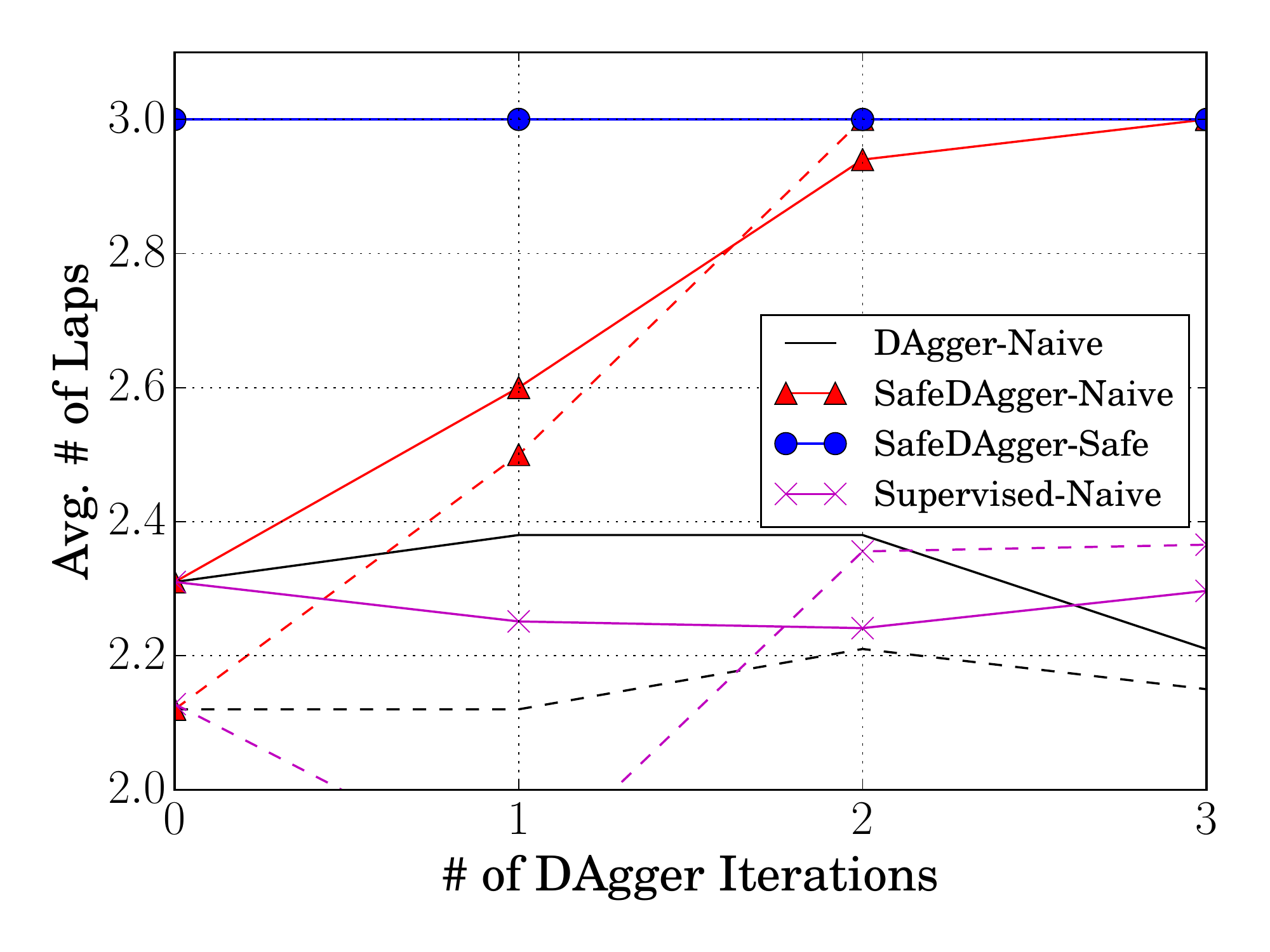}

        (a) 
    \end{minipage}
    \begin{minipage}{0.32\textwidth}
        \centering
        \includegraphics[width=\columnwidth]{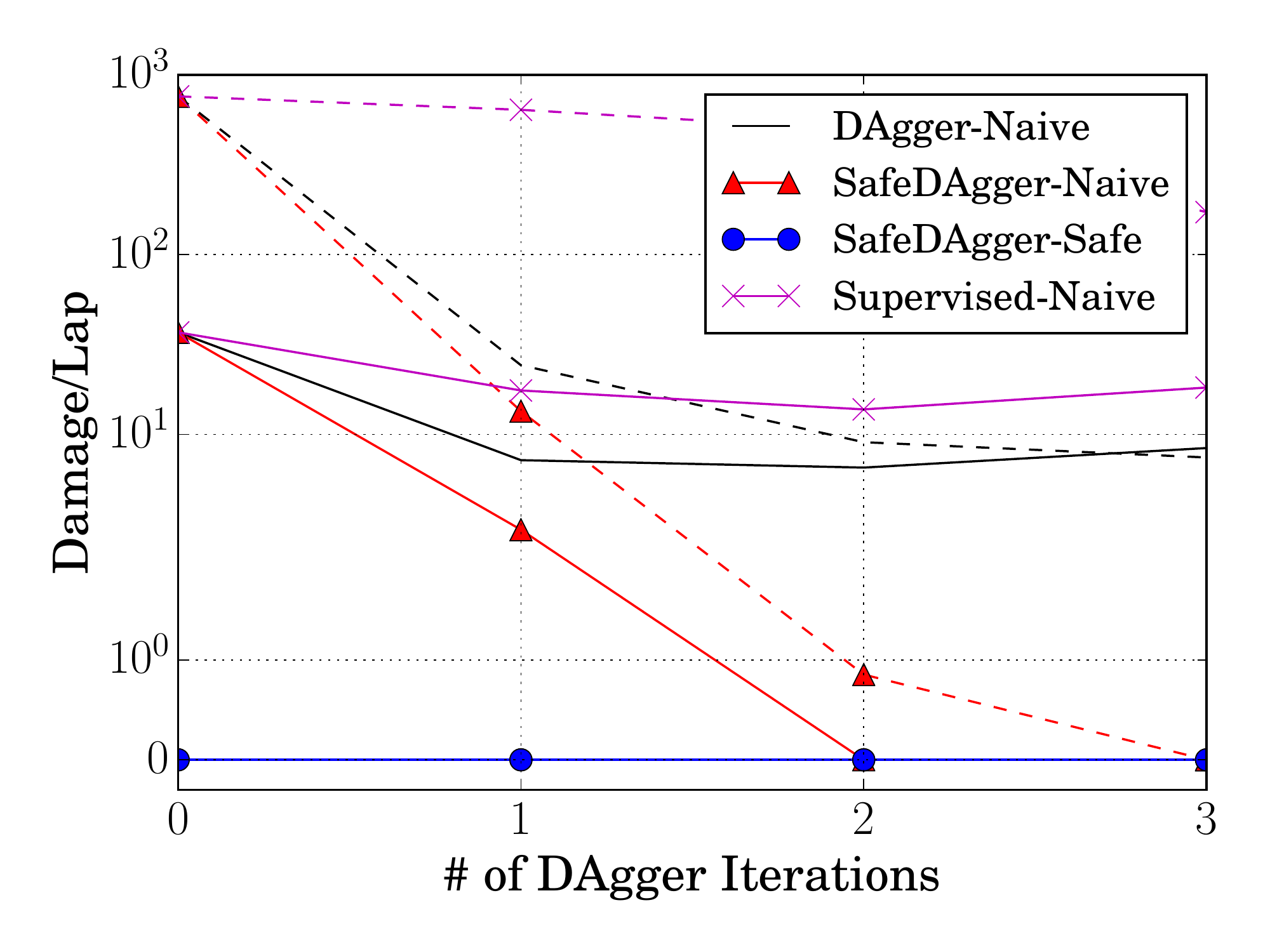}

        (b)
    \end{minipage}
    \begin{minipage}{0.32\textwidth}
        \centering
        \includegraphics[width=\columnwidth]{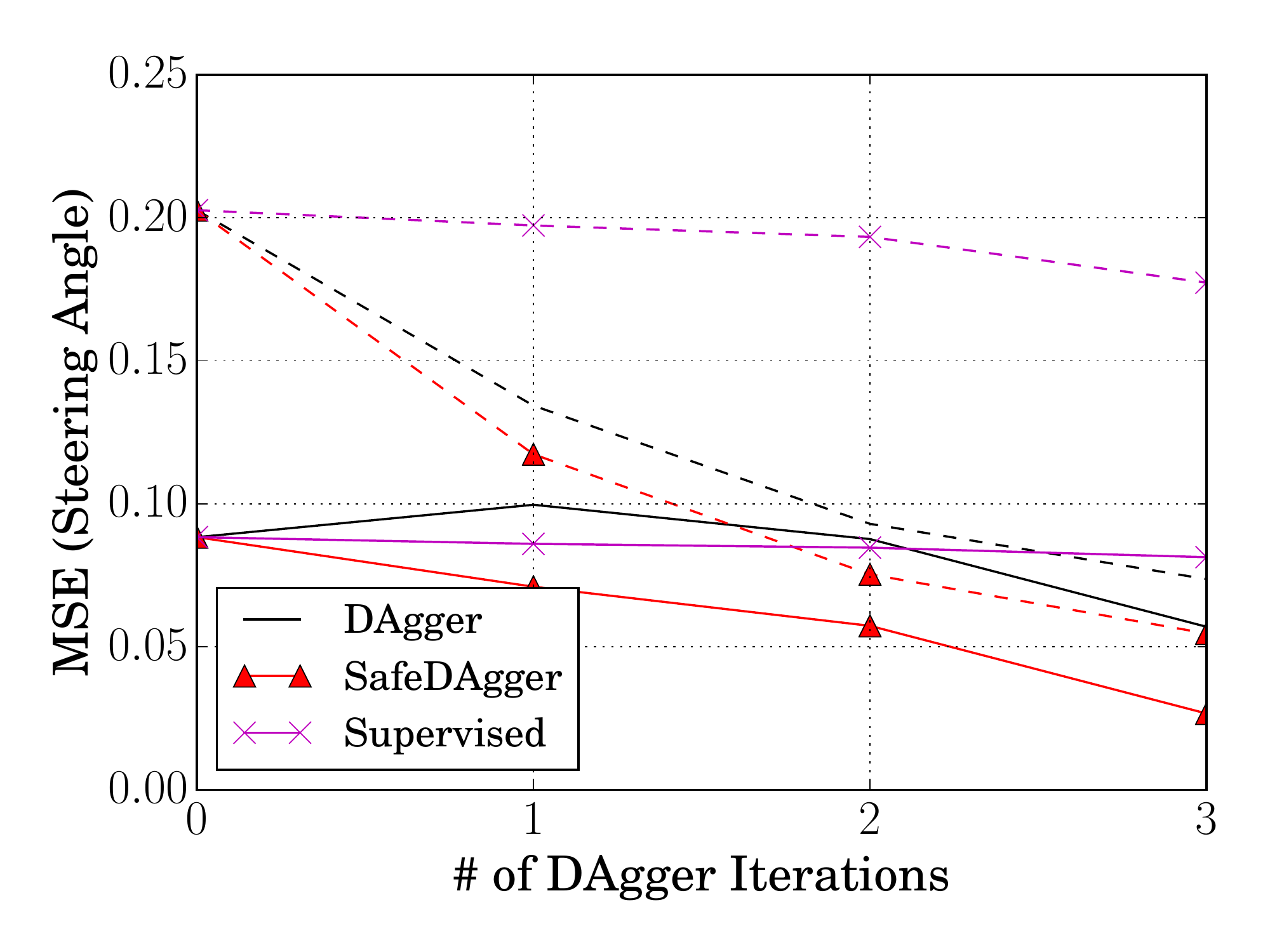}

        (c)
    \end{minipage}

        \caption{
            (a) Average number of laps ($\uparrow$), (b) damage per lap
            ($\downarrow$) and (c) the mean squared error of steering angle 
            for each configuration (training strategy--driving strategy) 
            over the iterations. We use solid and dashed curves for the cases 
            without and with traffic, respectively.
        }
        \label{fig:avgs}
    \vspace{-4mm}
\end{figure}

\begin{wrapfigure}{R}{0.32\textwidth}
    \centering
    \includegraphics[width=0.32\columnwidth]{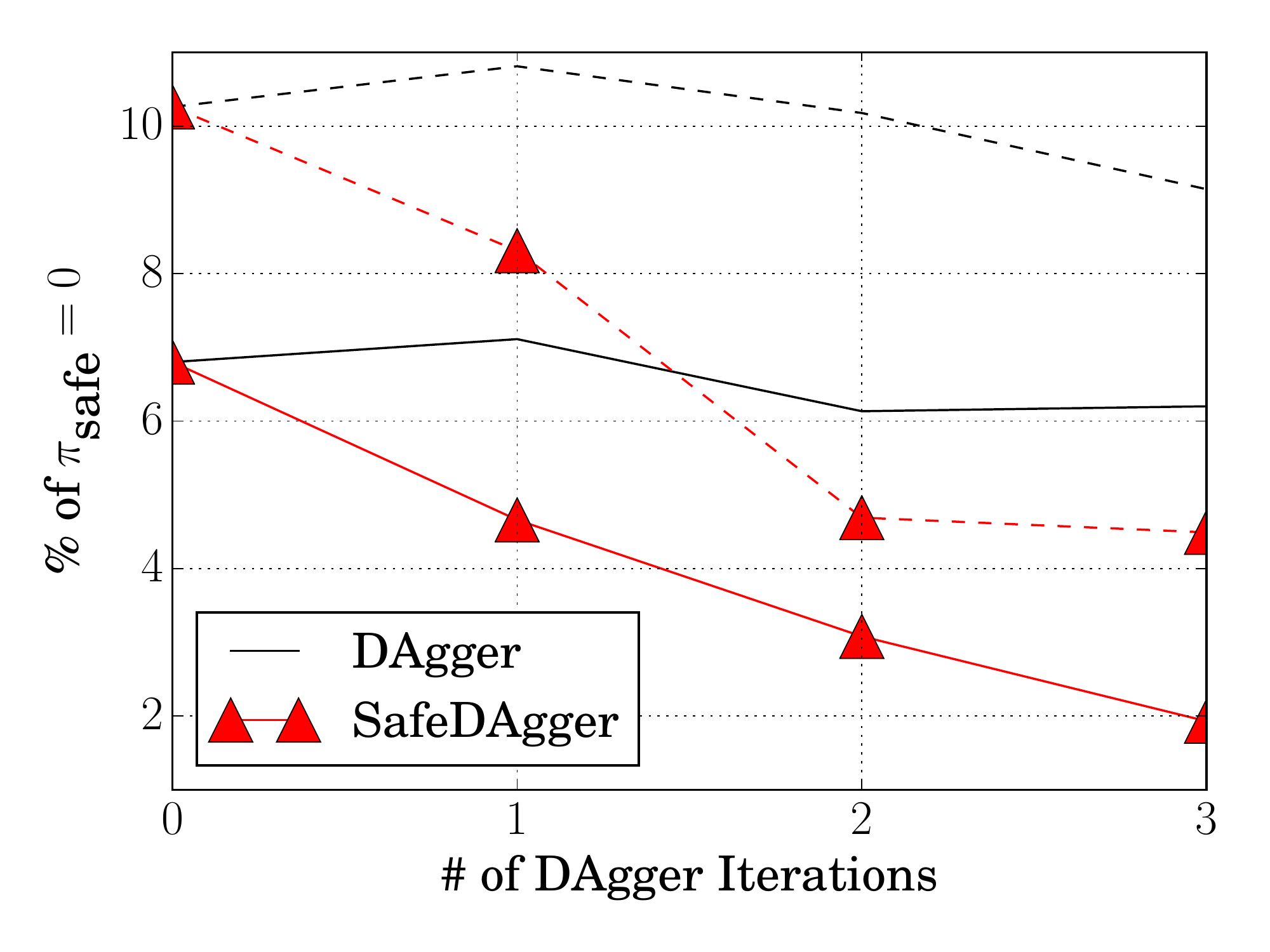}

    \caption{The portion of time driven by a reference policy during test. We
    see a clear downward trend as the iteration continues.}
    \label{fig:portion}
    \vspace{-5mm}
\end{wrapfigure}

In Fig.~\ref{fig:avgs}, we present the result in terms of both the average laps
and damage per lap. The first thing we notice is that a primary policy trained
using supervised learning (the 0-th iteration) alone works perfectly when a
safety policy is used together. The safety policy switched to the reference
policy for 7.11\% and 10.81\% of time without and with traffic during test.

Second, in terms of both metrics, the primary policy trained with the
proposed SafeDAgger makes much faster progress than the original DAgger.
After the third iteration, the primary policy trained with the SafeDAgger is
perfect. We conjecture that this is due to the effect of automated curriculum
learning of the SafeDAgger.  Furthermore, the examination of the mean squared
difference between the primary policy and the reference policy reveals that the
SafeDAgger more rapidly brings the primary policy closer to the reference
policy.

As a baseline we put the performance of a primary policy trained using purely
supervised learning in Fig.~\ref{fig:avgs}~(a)--(b). It clearly demonstrates
that supervised learning alone cannot train a primary policy well even when an
increasing amount of training examples are presented.

In Fig.~\ref{fig:portion}, we observe that the portion of time the safety policy
switches to the reference policy while driving decreases as the SafeDAgger
iteration progresses. We conjecture that this happens as the SafeDAgger
encourages the primary policy's learning to focus on those cases deemed
difficult by the safety policy. When the primary policy was trained with the
original DAgger (which does not take into account the difficulty of each
collected state), the rate of decrease was much smaller. Essentially, using the
safety policy and the SafeDAgger together results in a virtuous cycle of less
and less queries to the reference policy during both training and test.

Lastly, we conduct one additional run with the SafeDAgger while training a
safety policy to predict the deviation of a primary policy from the reference
policy one second in advance. We observe a similar trend, which makes the
SafeDAgger a realistic algorithm to be deployed in practice.

\section{Conclusion}

In this paper, we have proposed an extension of DAgger, called SafeDAgger. We
first introduced a safety policy which prevents a primary policy from falling
into a dangerous state by automatically switching between a reference policy and
the primary policy {\it without} querying the reference policy. This safety
policy is used during data collection stages in the proposed SafeDAgger, which
can collect a set of progressively difficult examples while minimizing the
number of queries to a reference policy.  The extensive experiments on simulated
autonomous driving showed that the SafeDAgger not only queries a reference
policy less but also trains a primary policy more efficiently.

Imitation learning, in the form of the SafeDAgger, allows a primary policy to
learn without any catastrophic experience. The quality of a learned policy is
however limited by that of a reference policy. More research in finetuning a
policy learned by the SafeDAgger to surpass existing, reference policies, for
instance by reinforcement learning \cite{silver2016mastering}, needs to be
pursued in the future. 

\subsubsection*{Acknowledgments}

We thank the support by Facebook, Google (Google Faculty Award 2016) and
NVidia (GPU Center of Excellence 2015-2016).

\bibliographystyle{abbrv}
\bibliography{safetynet}

\newpage
\appendix

\section{Dataset and Collection Procedure} 
\label{sec:dataset}

We use TORCS~\cite{torcs} to simulate autonomous driving in this paper.
The control frequency for driving the car in simulator is 30 Hz, 
sufficient enough for driving speed below 50 mph. 

\paragraph{Sensory Input}

\begin{figure}[h]
    \centering
    \includegraphics[width=\columnwidth]{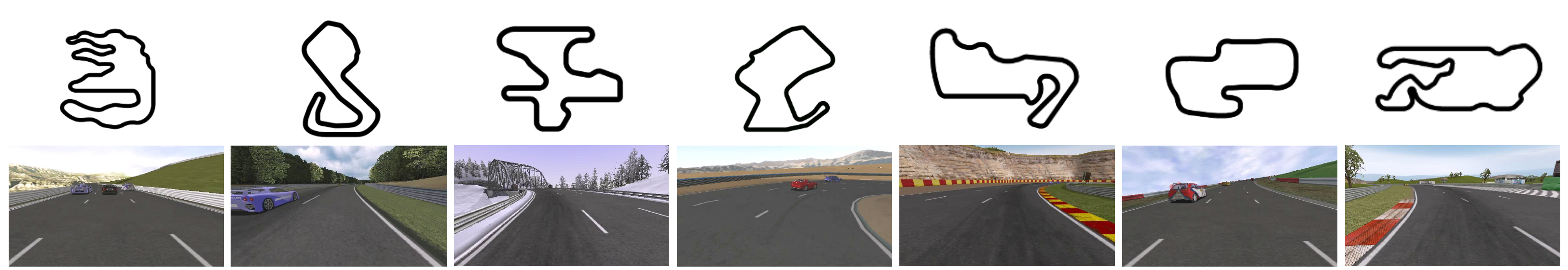}
    \\
    (a) Training tracks 

    \vspace{3mm}

    \includegraphics[width=\columnwidth]{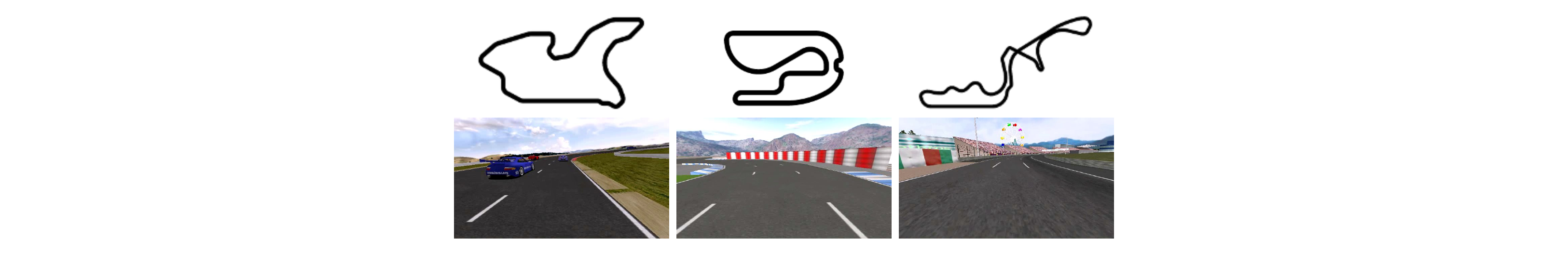}
    \\
    (b) Test tracks 

    \caption{Training and test tracks with sample frames.}
    \label{fig:tracks}
\end{figure}

We use a front-facing camera mounted on a racing car to collect image frames as
the car drives. Each image is scaled and cropped to $160\times72$ pixels with three
colour channels (R, G and B). In Fig.~\ref{fig:tracks}, we show the 
seven training tracks and three test tracks with one sample image frame per track.

\paragraph{Labels}

As the car drives, we collect the following twelve variables per image frame: 
\begin{enumerate}
    \item $I_{ll} \in \left\{0, 1\right\}$: if there is a lane to the left
    \item $I_{lr} \in \left\{0, 1\right\}$: if there is a lane to the right
    \item $I_{cl} \in \left\{0, 1\right\}$: if there is a car in front in the left lane
    \item $I_{cm} \in \left\{0, 1\right\}$: if there is a car in front in the same lane
    \item $I_{cr} \in \left\{0, 1\right\}$: if there is a car in front in the right lane
    \item $D_{cl} \in \RR$: distance to the car in front in the left lane
    \item $D_{cm} \in \RR$: distance to the car in front in the same lane 
    \item $D_{cr} \in \RR$: distance to the car in front in the right lane
    \item $P_c \in \left[ -1, 1\right]$: position of the car within the lane
    \item $A_c \in \left[ -1, 1\right]$: angle between the direction of the car and the direction of the lane
    \item \underline{$S_c \in \left[ -1, 1\right]$: angle of the steering wheel}
    \item \underline{$I_b \in \left\{0, 1\right\}$: if we brake the car}
\end{enumerate}
The first ten are state configurations that are observed only by a reference
policy but hidden to a primary policy and safety policy. The last two variables
are the control variables.  All the variables are used as target labels during
training, but only the last two ($S_c$ and $I_b$) are used during test to drive
a car.

\section{Policy Networks and Training}
\label{sec:policy}

\paragraph{Primary Policy Network}

We use a deep convolutional network that has five convolutional layers followed
by a group of fully-connected layers. In Table~\ref{table:modelConf}, we detail
the configuration of the network. 

\begin{table}[t]
    \small
    \centering
    \begin{tabular}{|c |  c  | c  | c  | c  | c||c||c  | c  | c  | c  | c|}
        \hline
        \multicolumn{12}{|c|}{\cellcolor{gray!10} Input - 3$\times$160$\times$72} \\
        \hline
        \hline
        \multicolumn{12}{|c|}{\cellcolor{gray!10} Conv1 - 64$\times$3$\times$3}\\
        \hline
        \hline
        \multicolumn{12}{|c|}{\cellcolor{gray!10} Max Pooling - $2 \times 2$}\\
        \hline
        \hline
        \multicolumn{12}{|c|}{\cellcolor{gray!10} Conv2 - 64$\times$3$\times$3}\\
        \hline
        \hline
        \multicolumn{12}{|c|}{\cellcolor{gray!10} Max Pooling - $2\times 2$}\\
        \hline
        \hline
        \multicolumn{12}{|c|}{\cellcolor{gray!10} Conv3 - 64$\times$3$\times$3}\\
        \hline
        \hline
        \multicolumn{12}{|c|}{\cellcolor{gray!10} Max Pooling - $2\times 2$}\\
        \hline
        \hline
        \multicolumn{12}{|c|}{\cellcolor{gray!10} Conv4 - 64$\times$3$\times$3}\\
        \hline
        \hline
        \multicolumn{12}{|c|}{\cellcolor{gray!10} Max Pooling - $2\times 2$}\\
        \hline
        \hline
        \multicolumn{12}{|c|}{\cellcolor{gray!10} \underline{Conv5} - 128$\times$5$\times$5}\\
        \hline
        \hline
        \multirow{2}{*}{fc-2} & 
        \multirow{2}{*}{fc-2} & 
        \multirow{2}{*}{fc-2} & 
        \multirow{2}{*}{fc-2} & 
        \multirow{2}{*}{fc-2} & 
        \cellcolor{gray!10}  & 
        \cellcolor{gray!10} fc-64 & 
        \multirow{2}{*}{fc-1} & 
        \multirow{2}{*}{fc-1} & 
        \multirow{2}{*}{fc-1} & 
        \multirow{2}{*}{fc-1} & 
        \multirow{2}{*}{fc-1} 
        \\
        & & & & &
        \multirow{-2}{*}{\cellcolor{gray!10} fc-2}
        &
        \cellcolor{gray!10} fc-1 & 
        & & & & \\
        \hline
        \hline
        $I_{ll}$ & 
        $I_{lr}$ & 
        $I_{cl}$ & 
        $I_{cm}$ & 
        $I_{cr}$ & 
        \cellcolor{gray!10} \underline{$I_b$} &
        \cellcolor{gray!10} \underline{$S_c$} & 
        $D_{cl}$ &
        $D_{cm}$ &
        $D_{cr}$ &
        $P_c$ & 
        $A_c$  \\
        \hline
    \end{tabular}

    \captionof{figure}{
        The configuration of a primary policy network. Each convolutional layer
        is denoted by ``Conv - \# channels $\times$ height $\times$ width''.
        Max pooling without overlap follows each convolutional layer. We use
        rectified linear units~\cite{nair2010rectified,glorot2011deep} for
        point-wise nonlinearities. Only the shaded part of the full network is
        used during test.
    }
    \label{table:modelConf}
\end{table}

\paragraph{Safety Policy Network}

A feedforward network with two fully-connected hidden layers of rectified
linear units is used to implement a safety policy. This safety policy network
takes as input the activations of {\it \underline{Conv5}} of the primary policy
network (see Fig.~\ref{table:modelConf}.)

\paragraph{Training}

Given a set of training examples, we use stochastic gradient descent (SGD) with
a batch size of 64, momentum of 0.9, weight decay of 0.001 and initial
learning rate of 0.001 to train a policy network. During training, the learning
rate is divided by 5 each time the validation error stops improving. When the
validation error increases, we early-stop the training. In most cases,
training takes approximately 40 epochs.

\section{Sample Image Frames}

    \newcommand{\safeframe}[1]{\includegraphics[width=0.19\linewidth]{frames/safe_frames_#1_compress.pdf}\hfill}
    \newcommand{\unsafeframe}[1]{\includegraphics[width=0.19\linewidth]{frames/unsafe_frames_#1_compress.pdf}\hfill}

\begin{figure}[t]
    \begin{minipage}{\linewidth}
        \centering
        {\bf Safe Frames}

        \safeframe{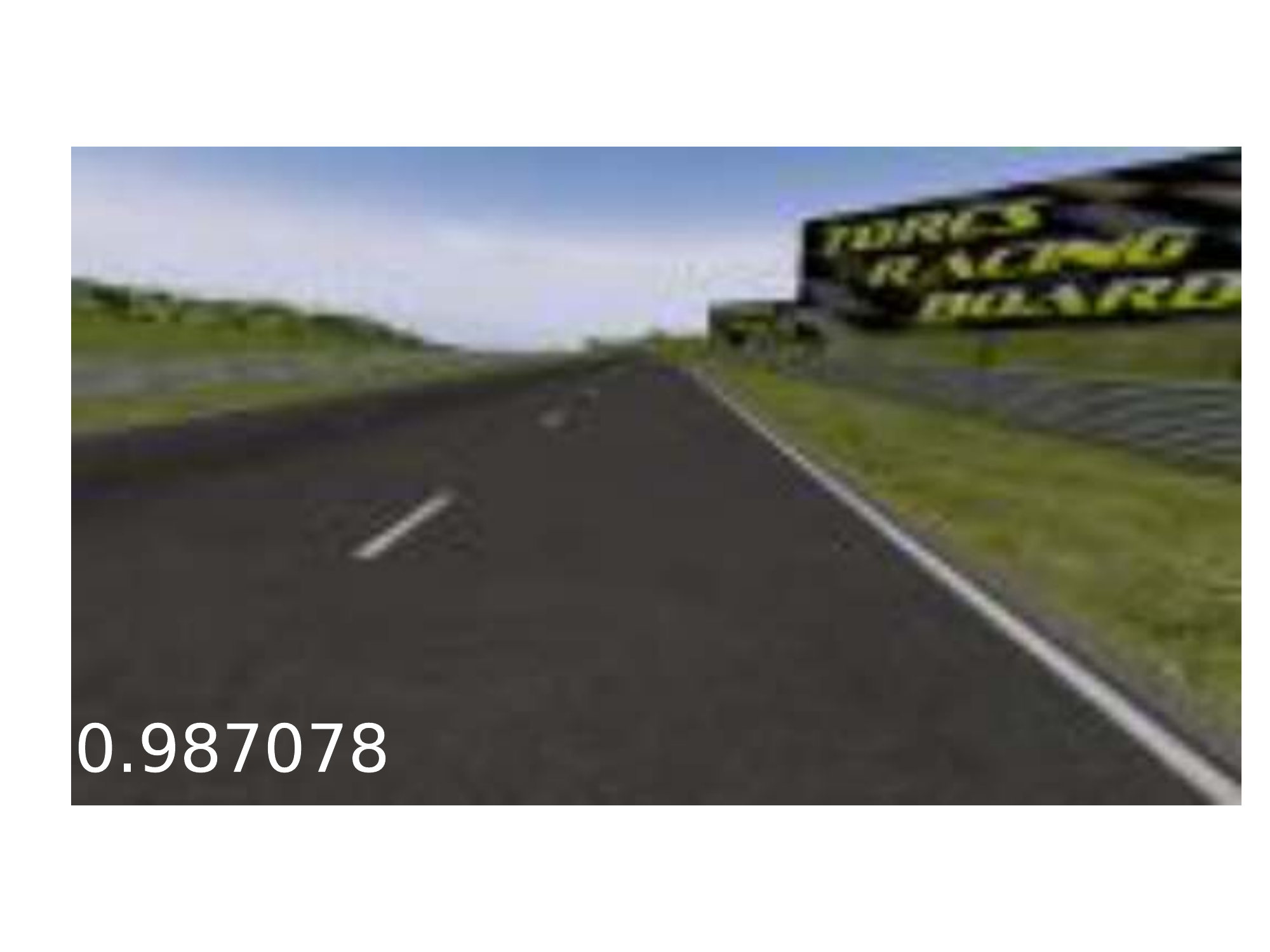}
        \safeframe{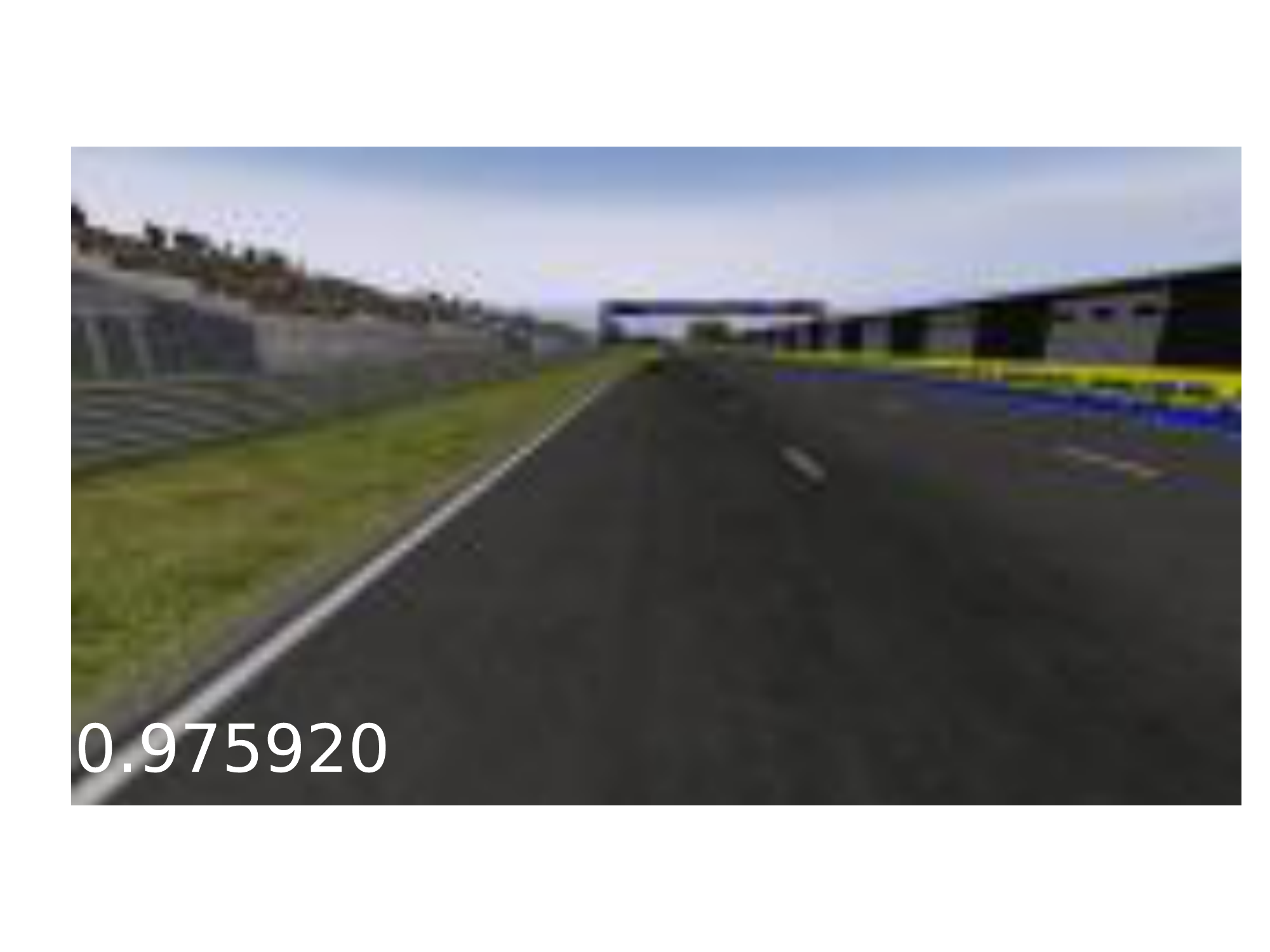}
        \safeframe{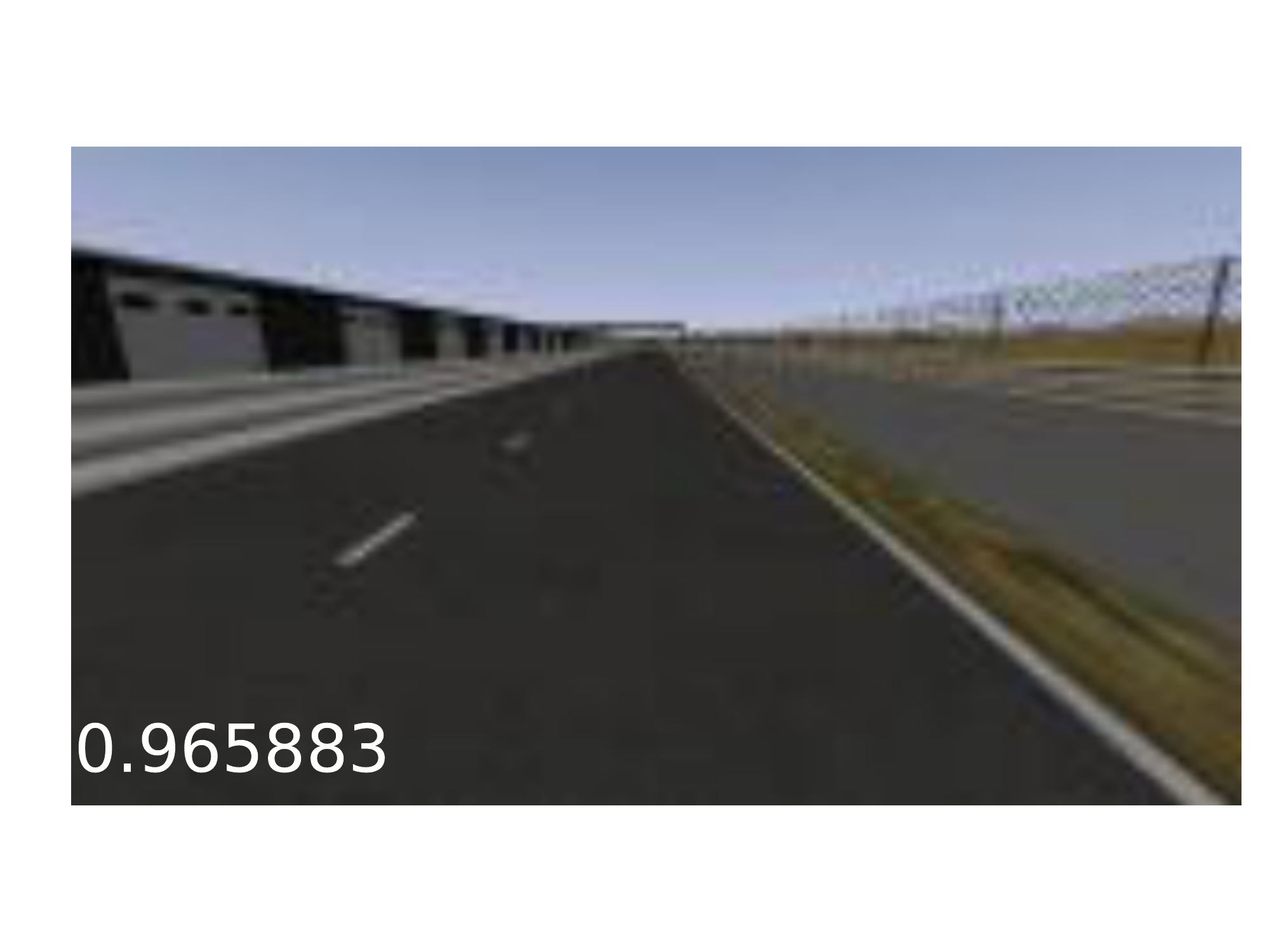}
        \safeframe{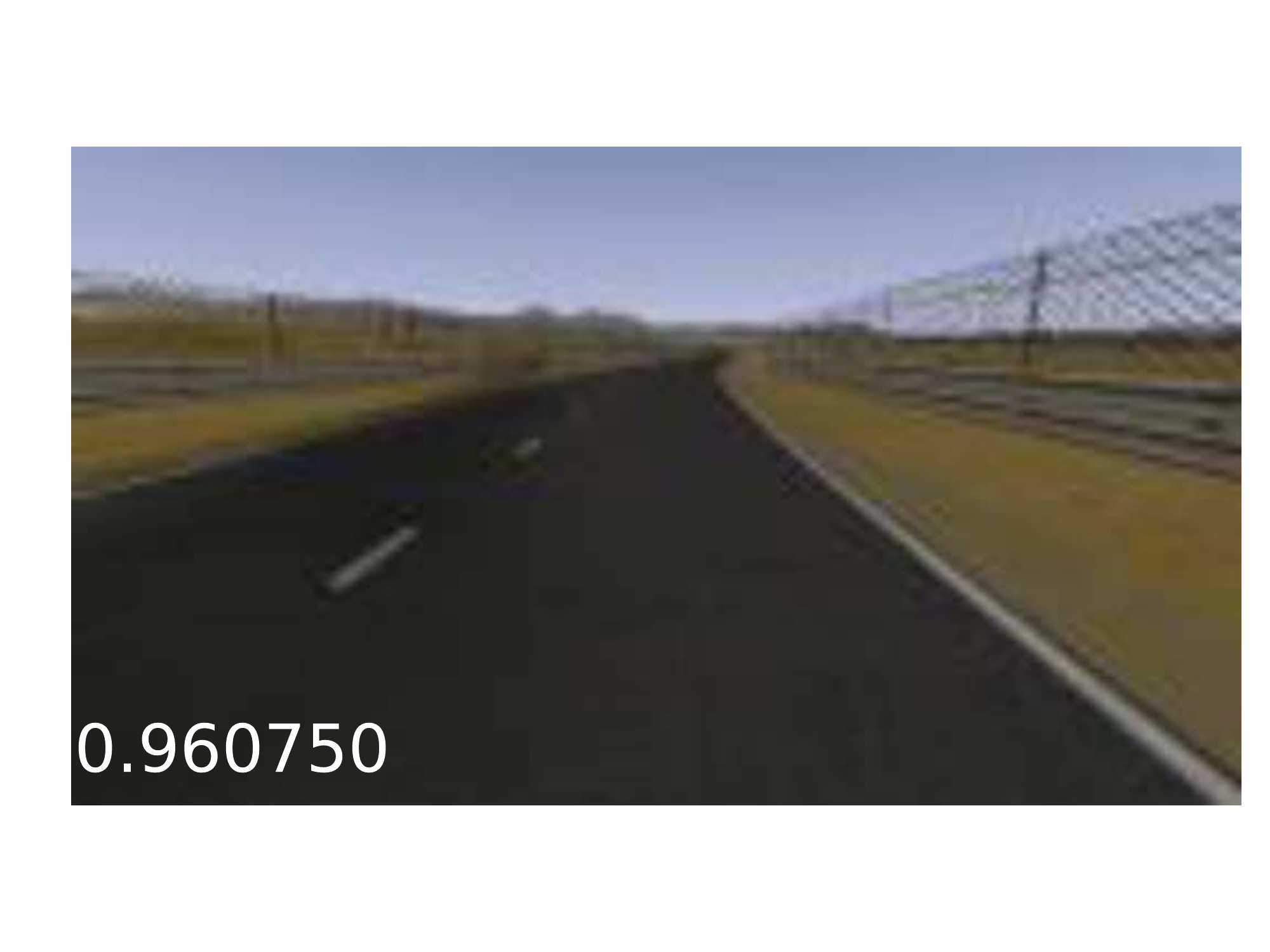}
        \safeframe{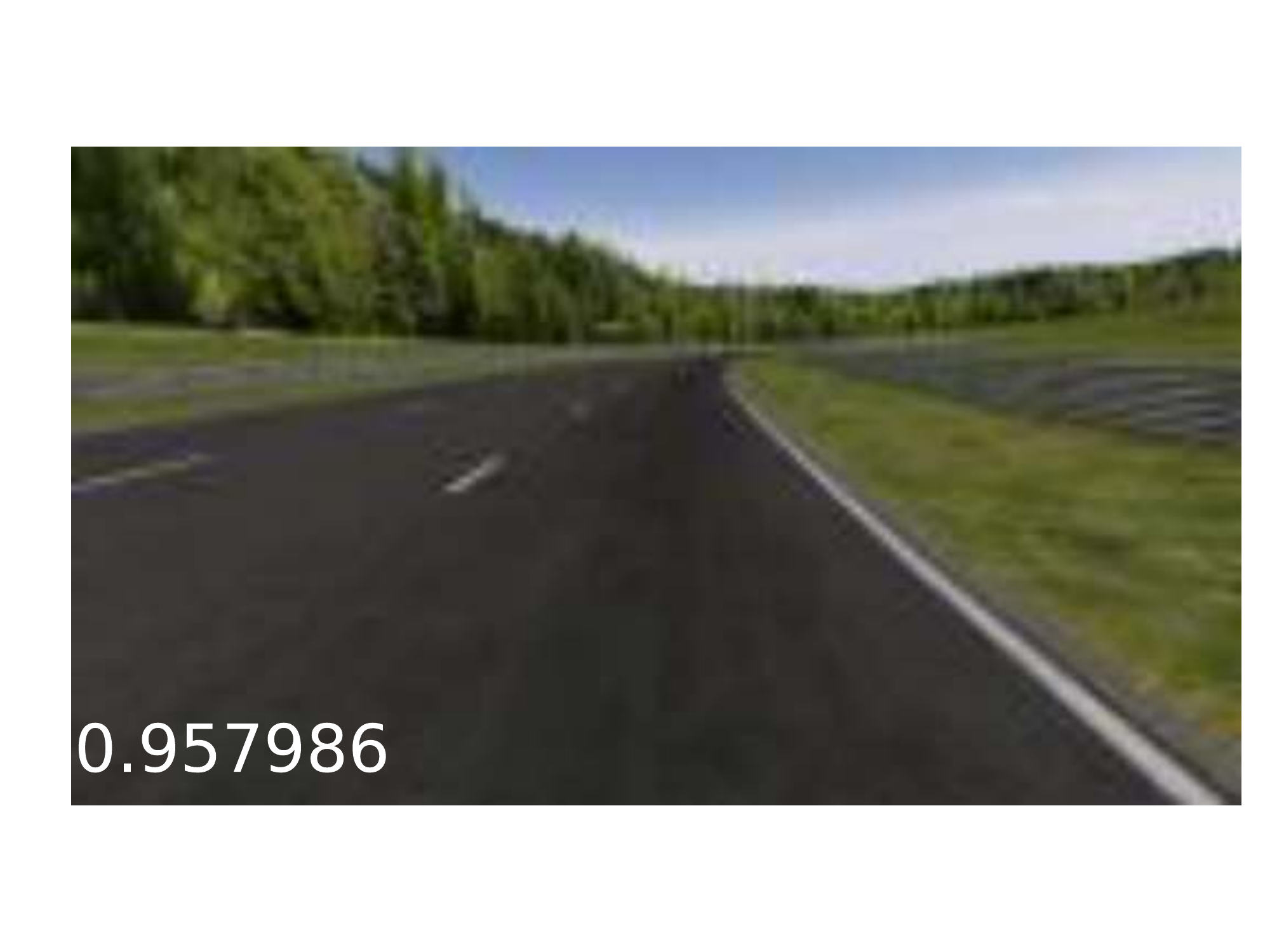}

        \vspace{-3mm}
        \safeframe{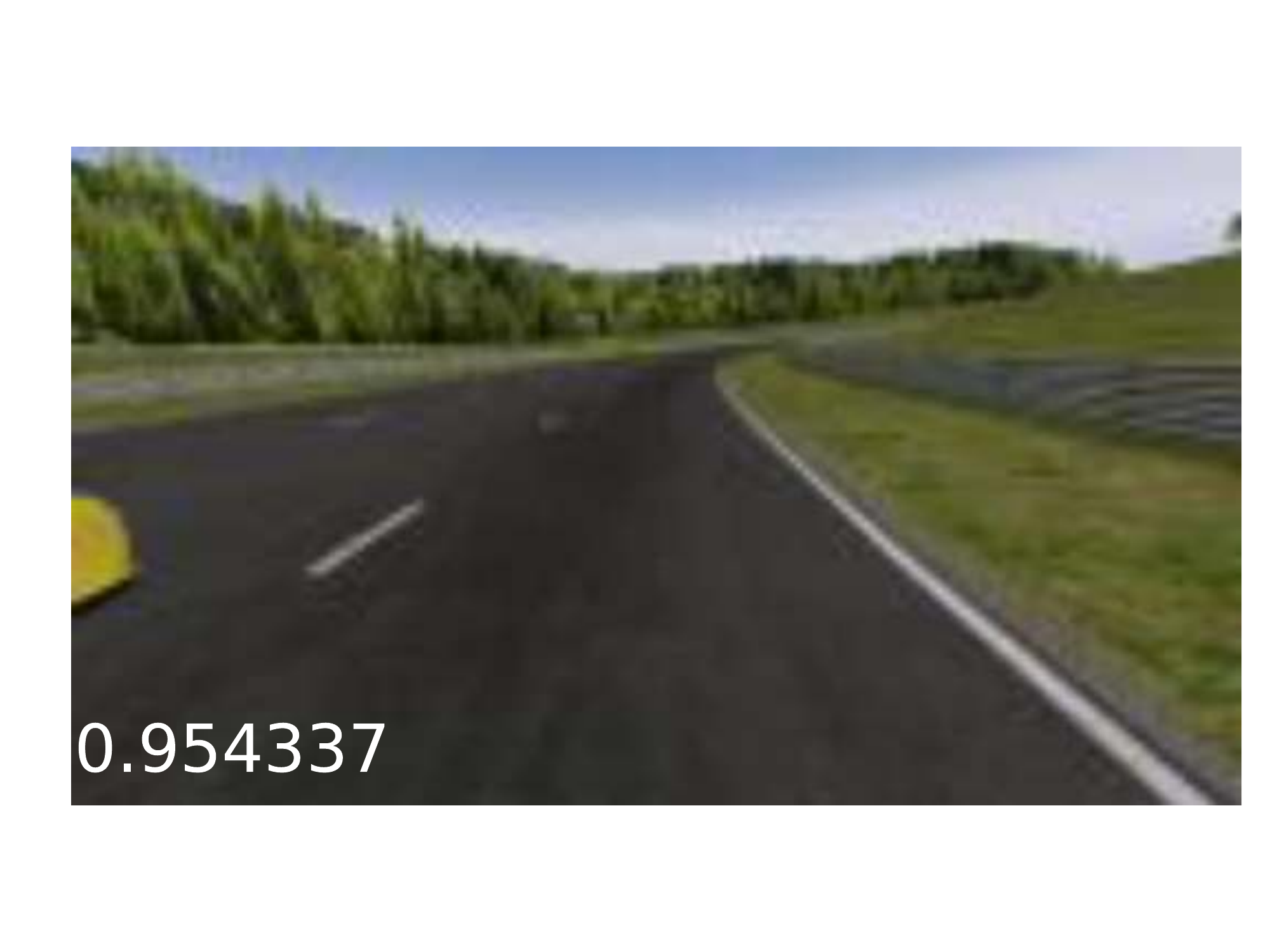}
        \safeframe{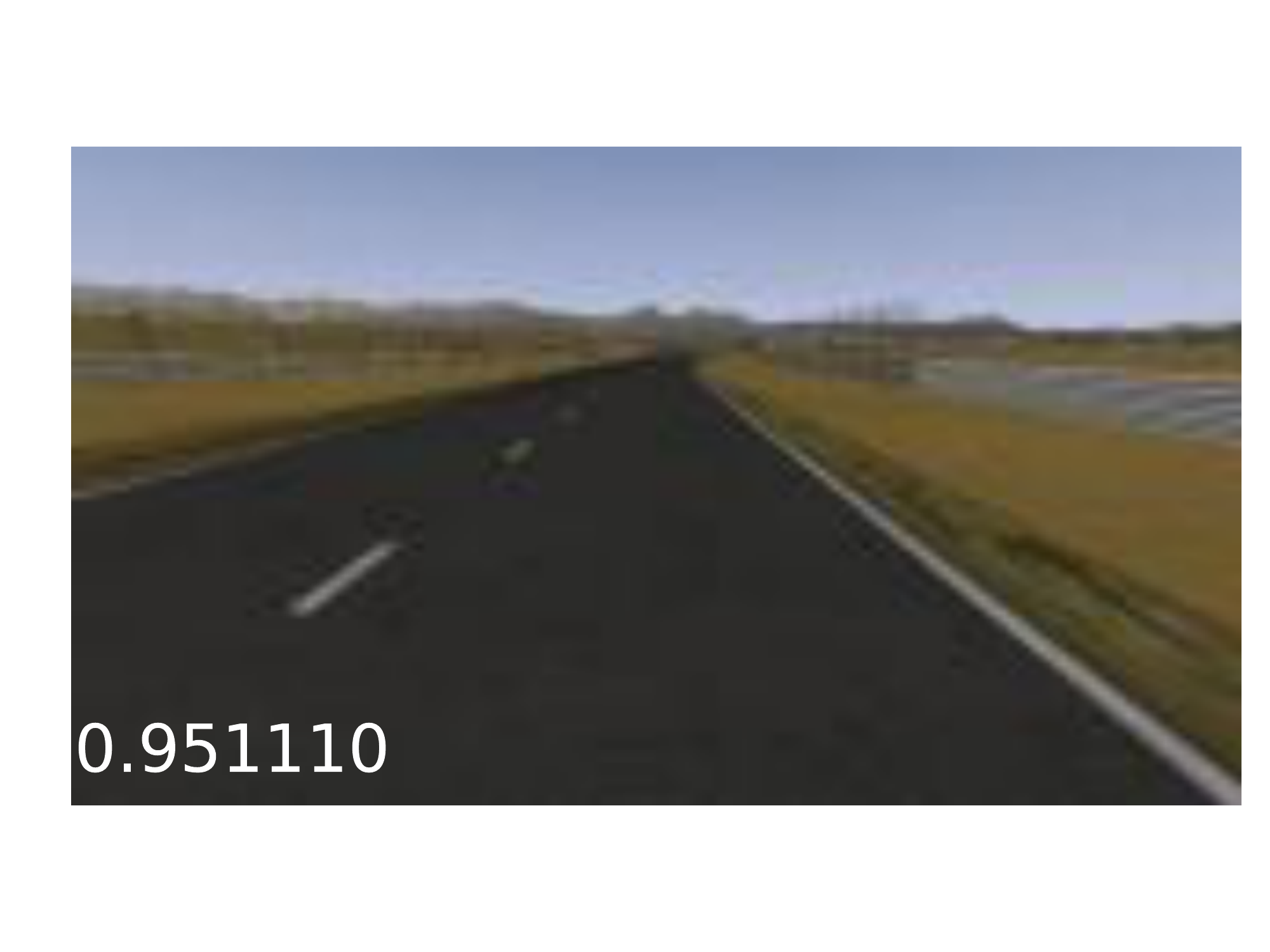}
        \safeframe{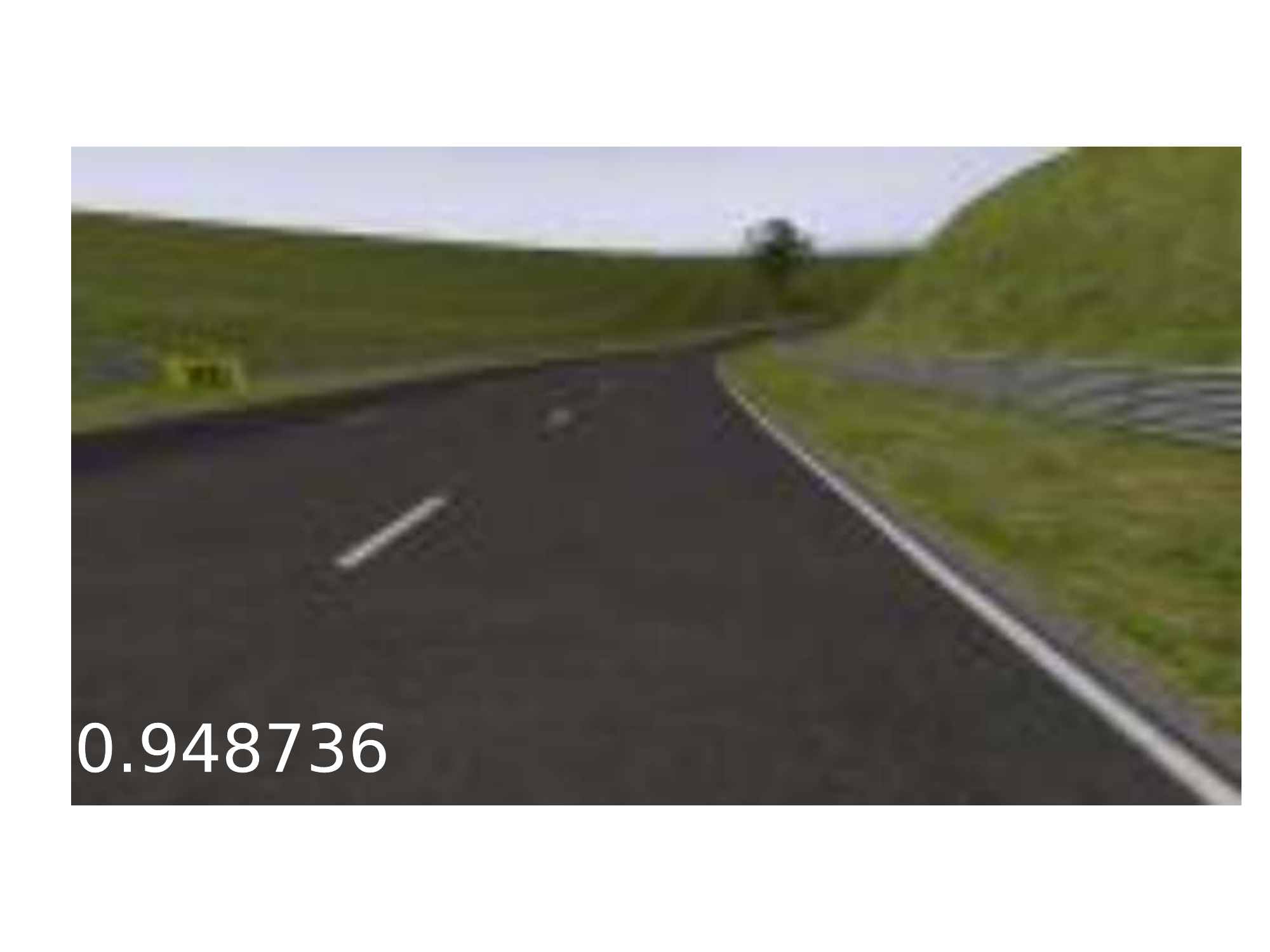}
        \safeframe{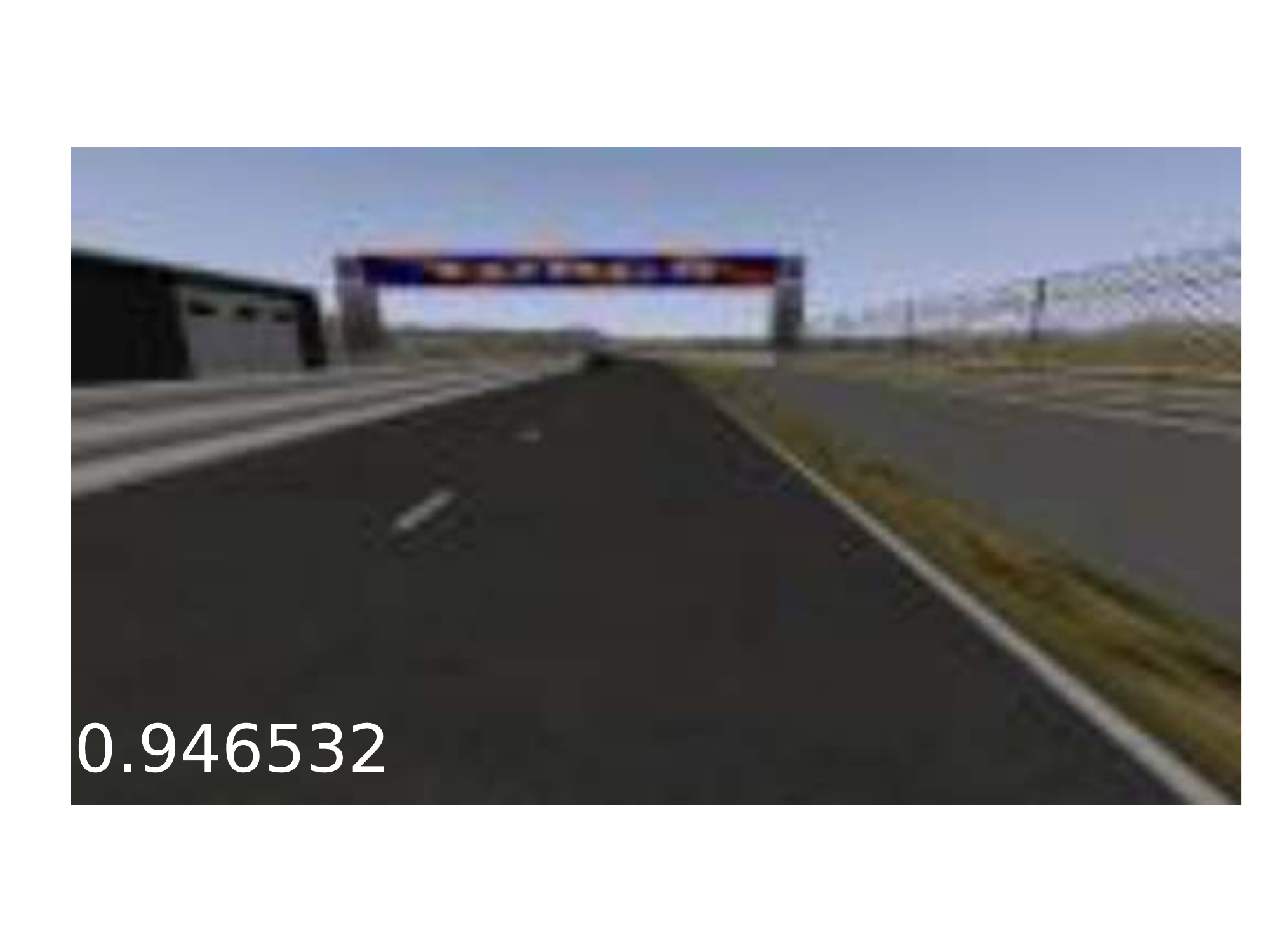}
        \safeframe{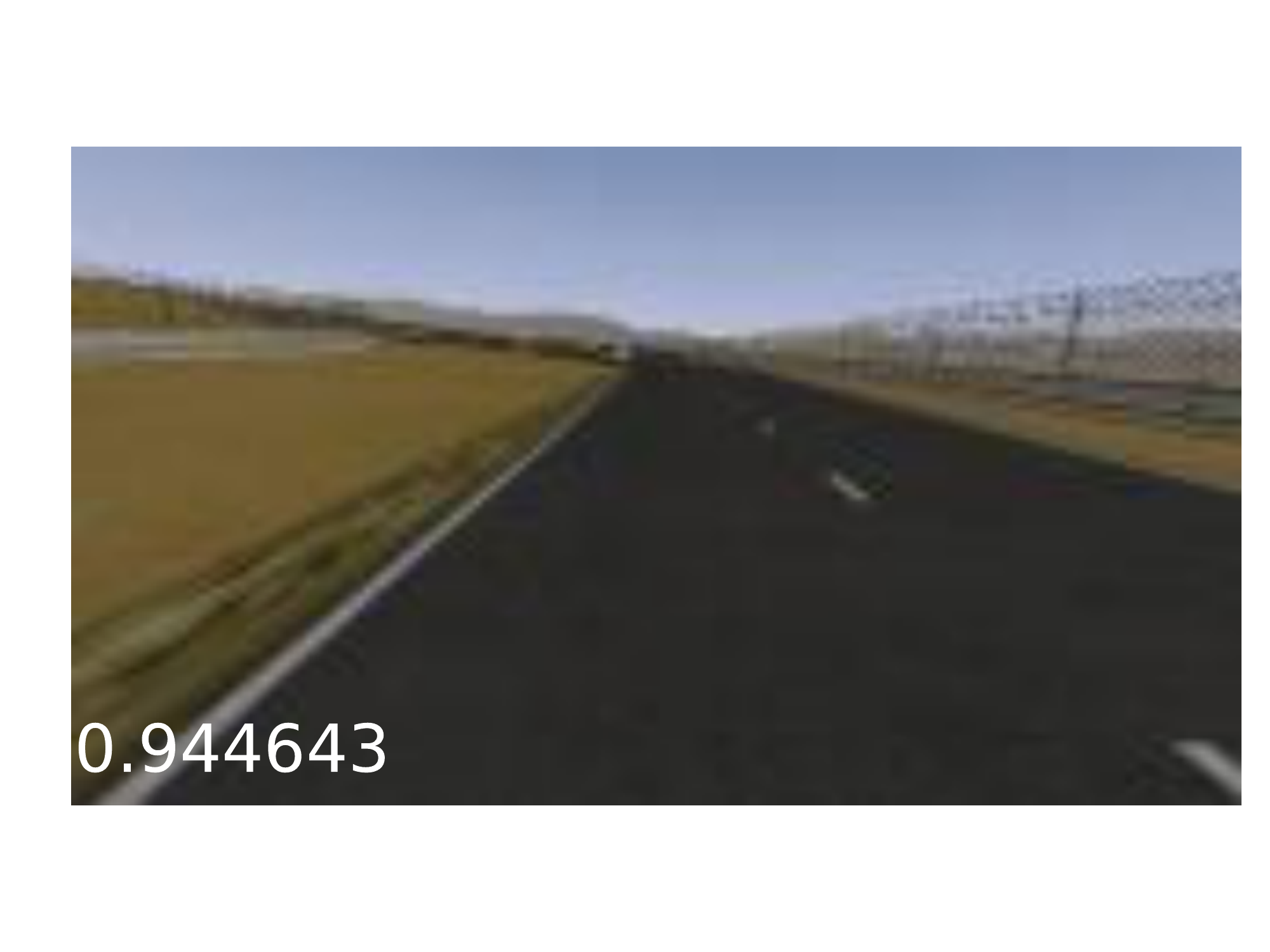}
    \end{minipage}

    \begin{minipage}{\linewidth}
        \centering
        {\bf Unsafe Frames}

        \unsafeframe{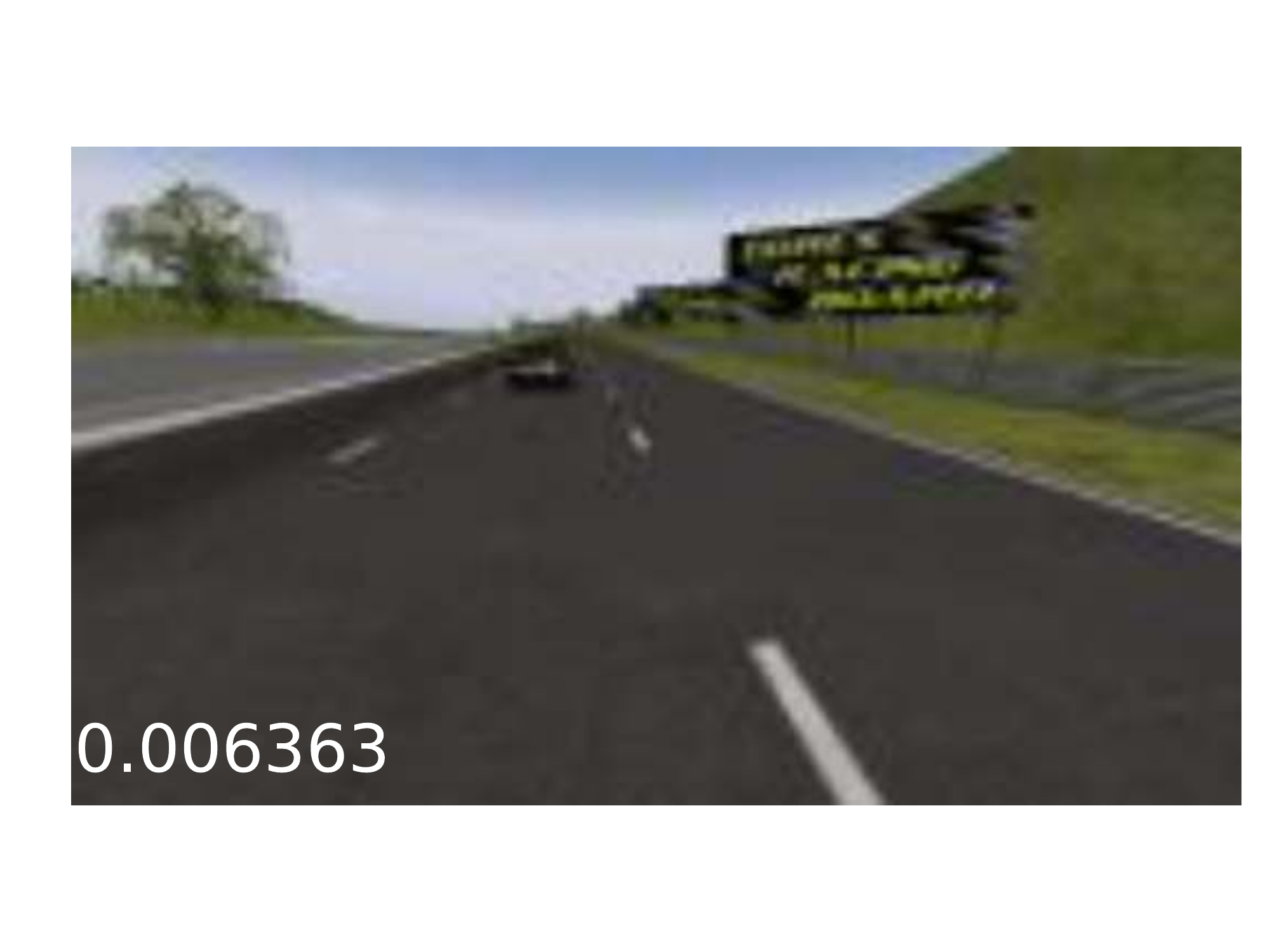}
        \unsafeframe{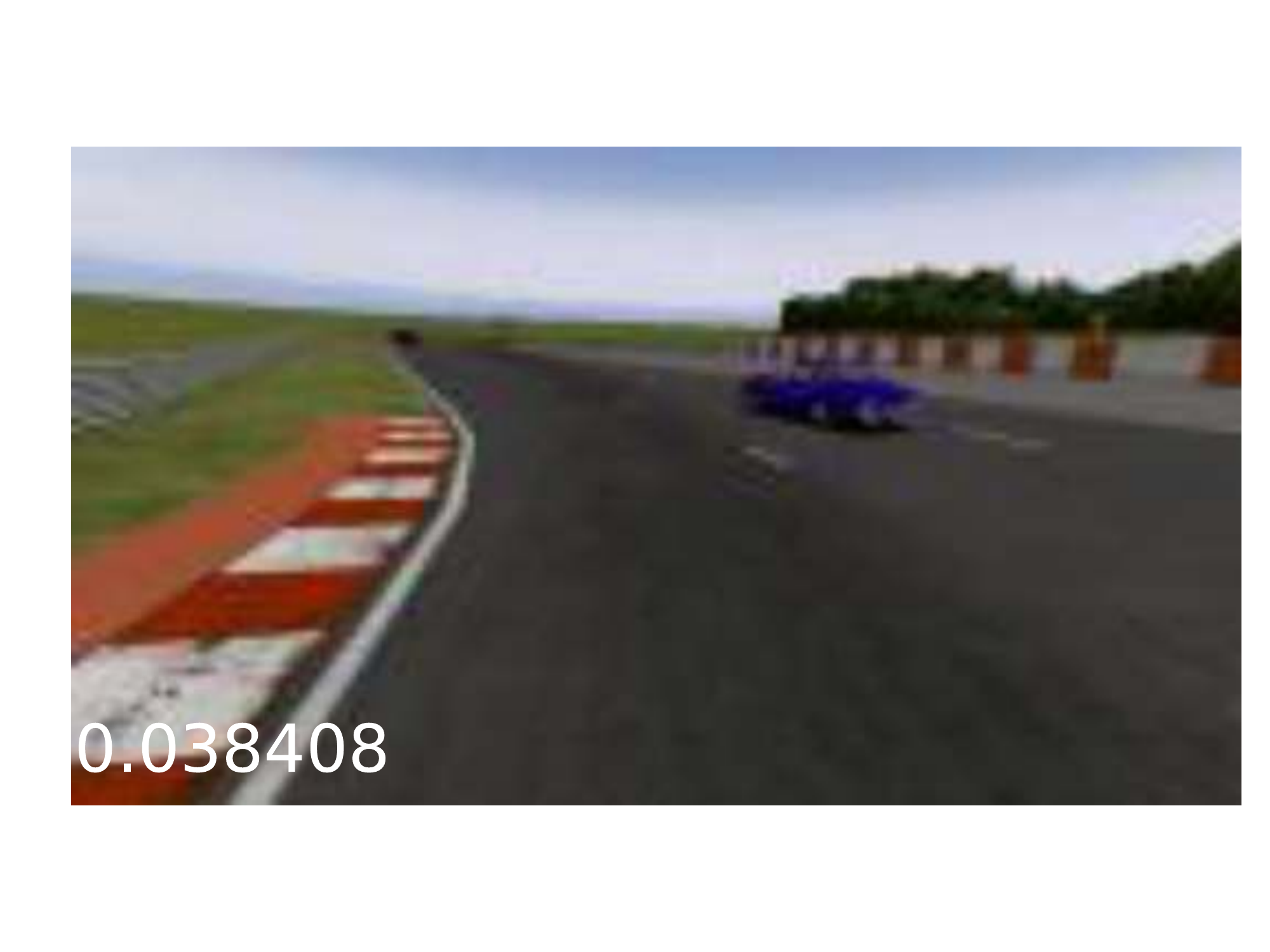}
        \unsafeframe{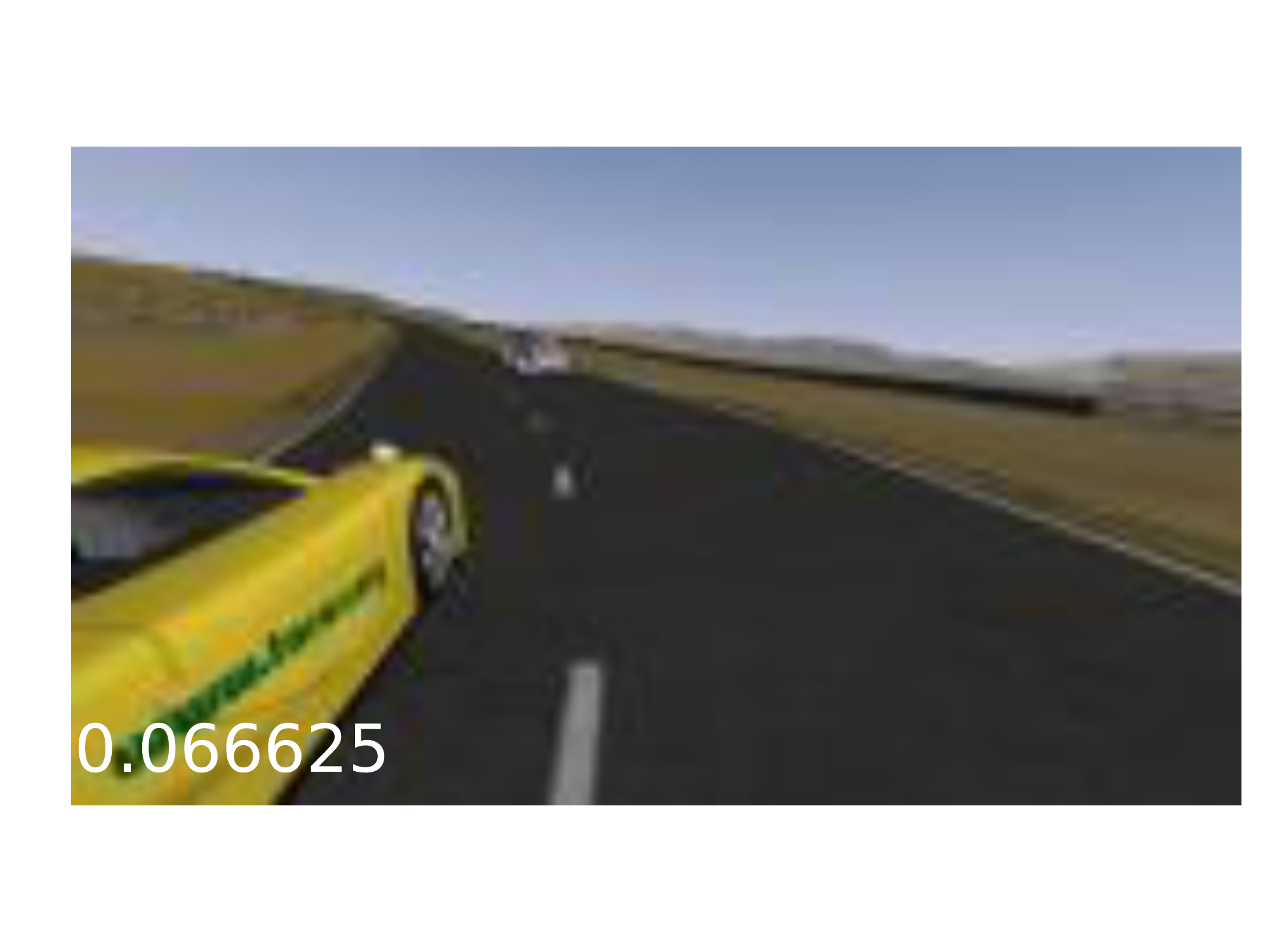}
        \unsafeframe{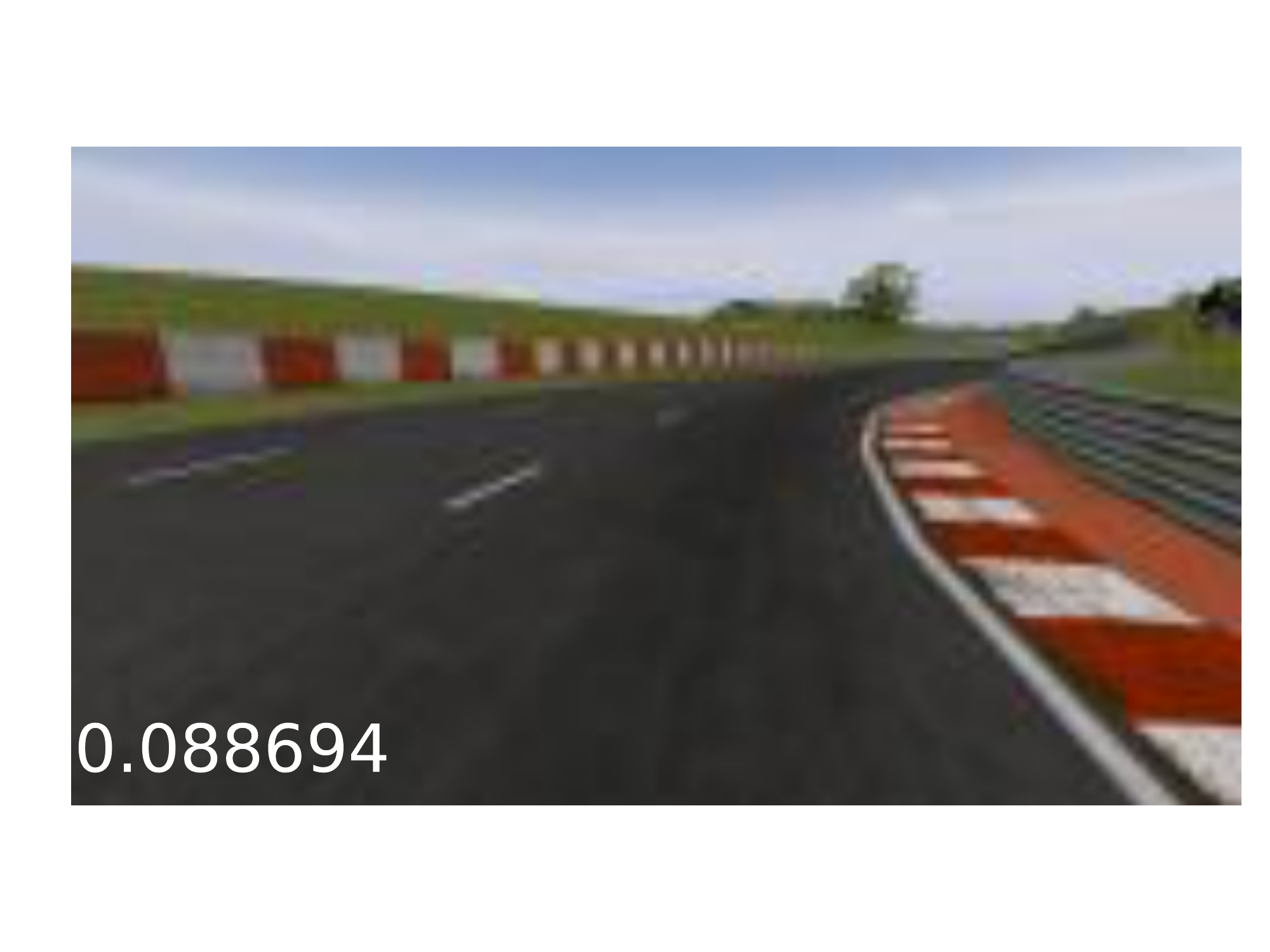}
        \unsafeframe{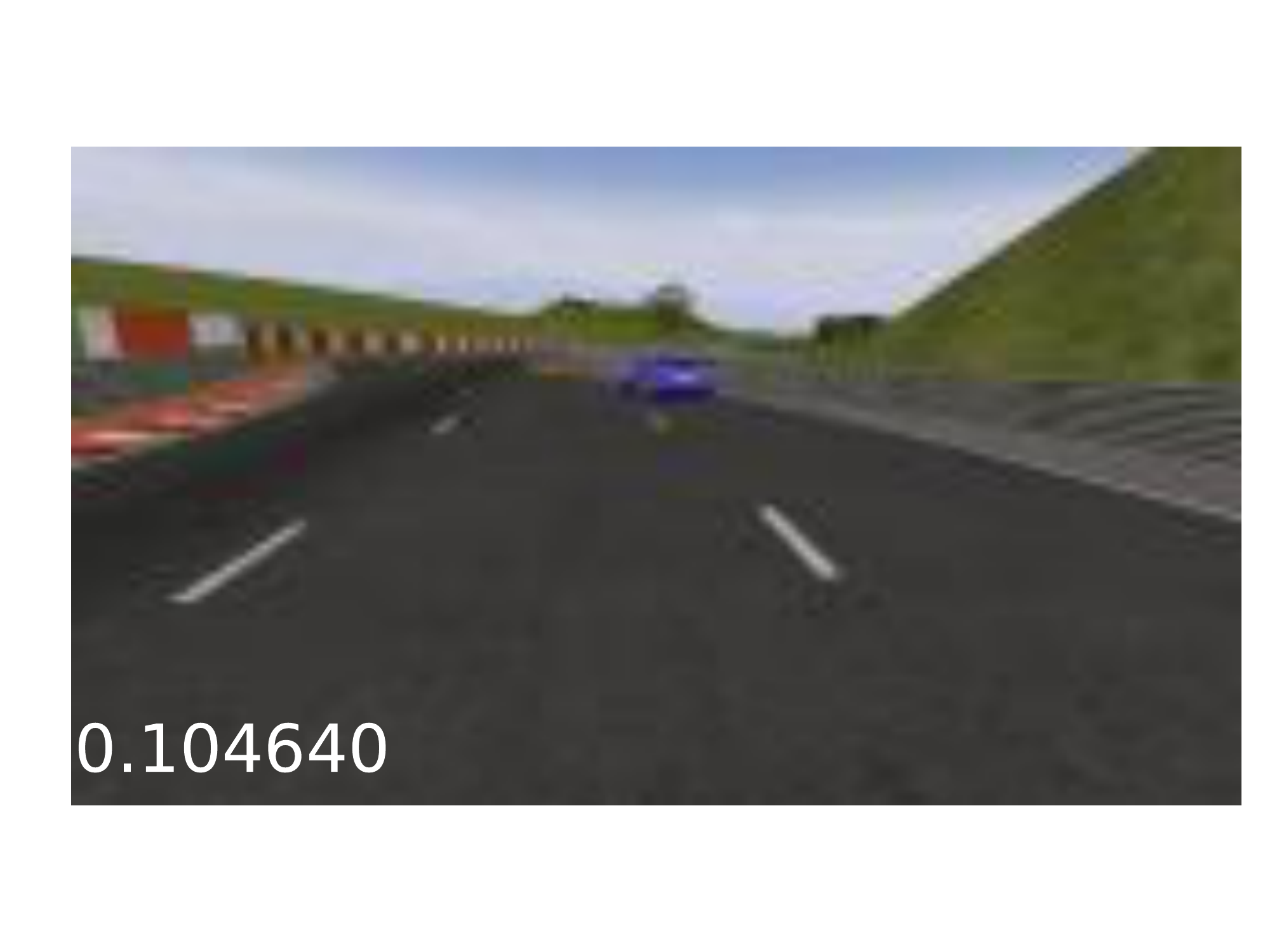}

        \vspace{-3mm}
        \unsafeframe{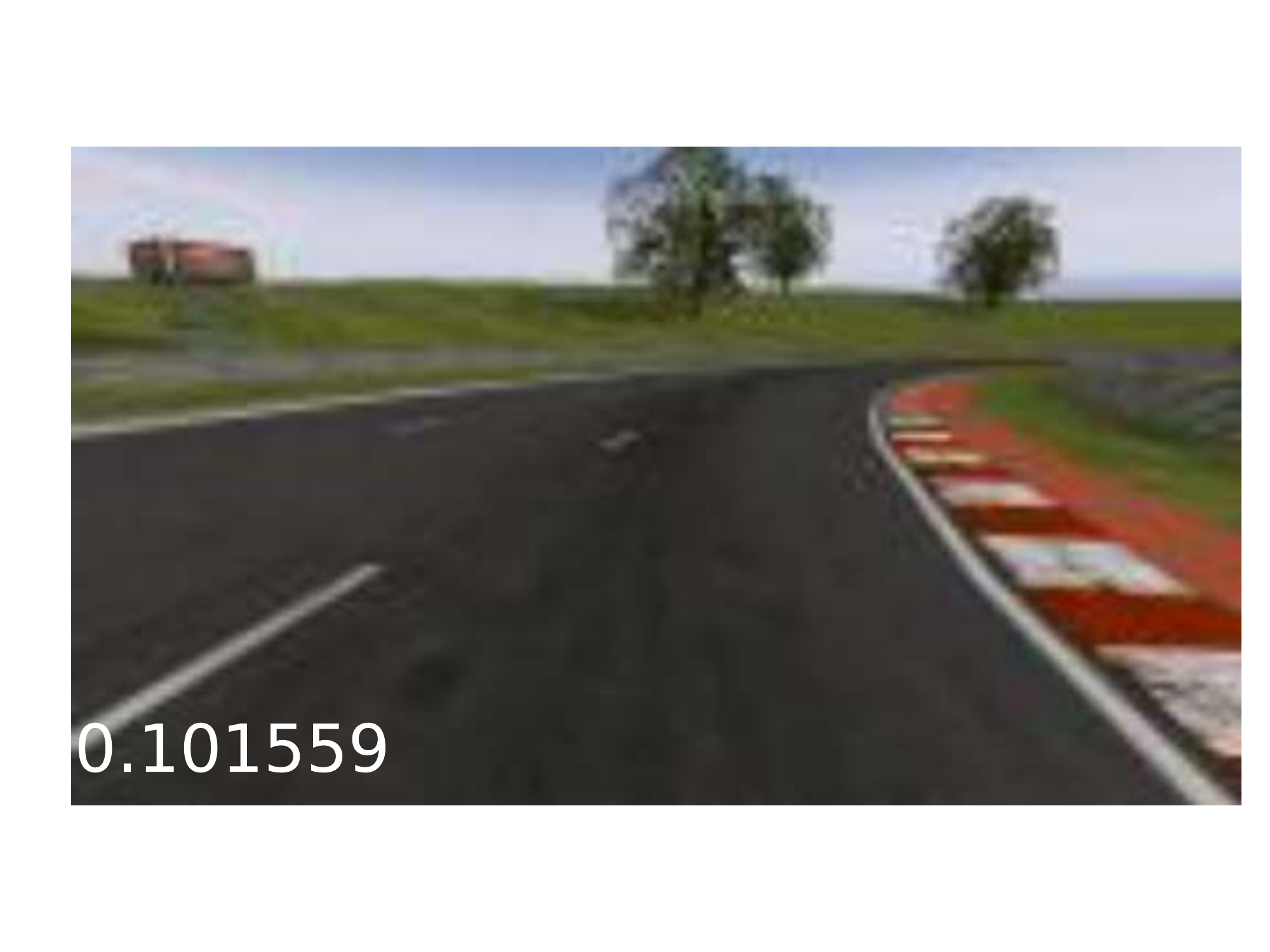}
        \unsafeframe{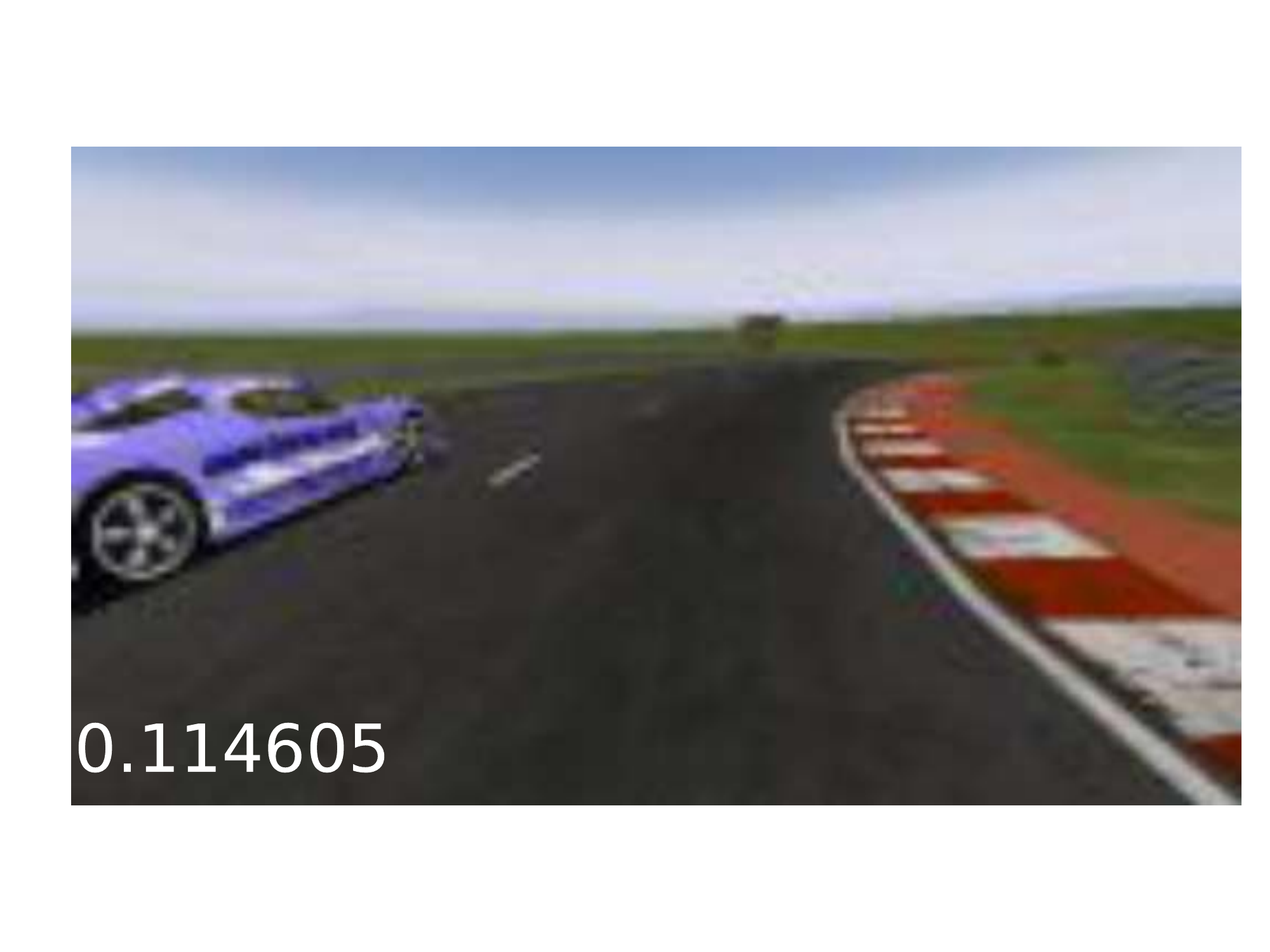}
        \unsafeframe{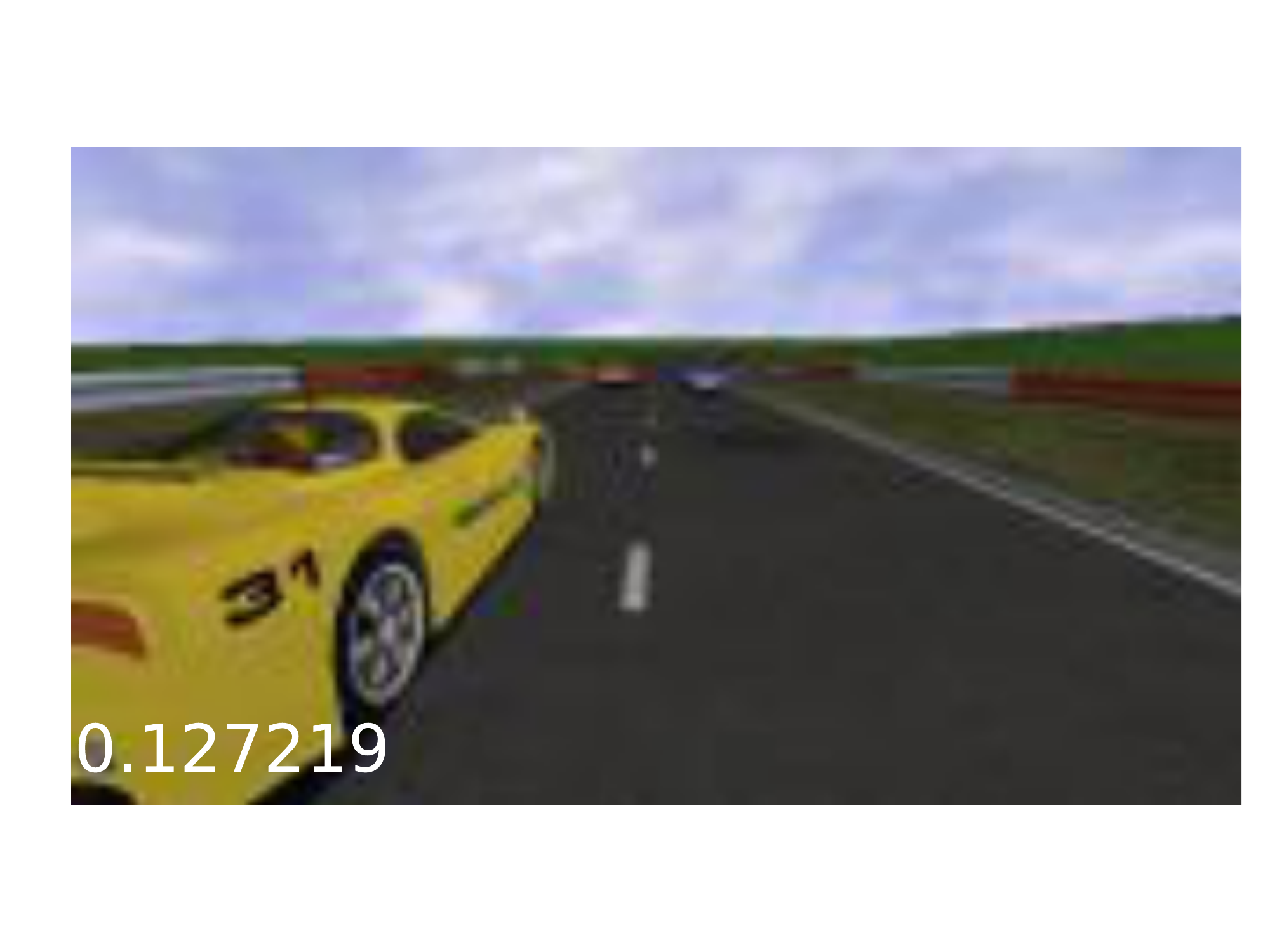}
        \unsafeframe{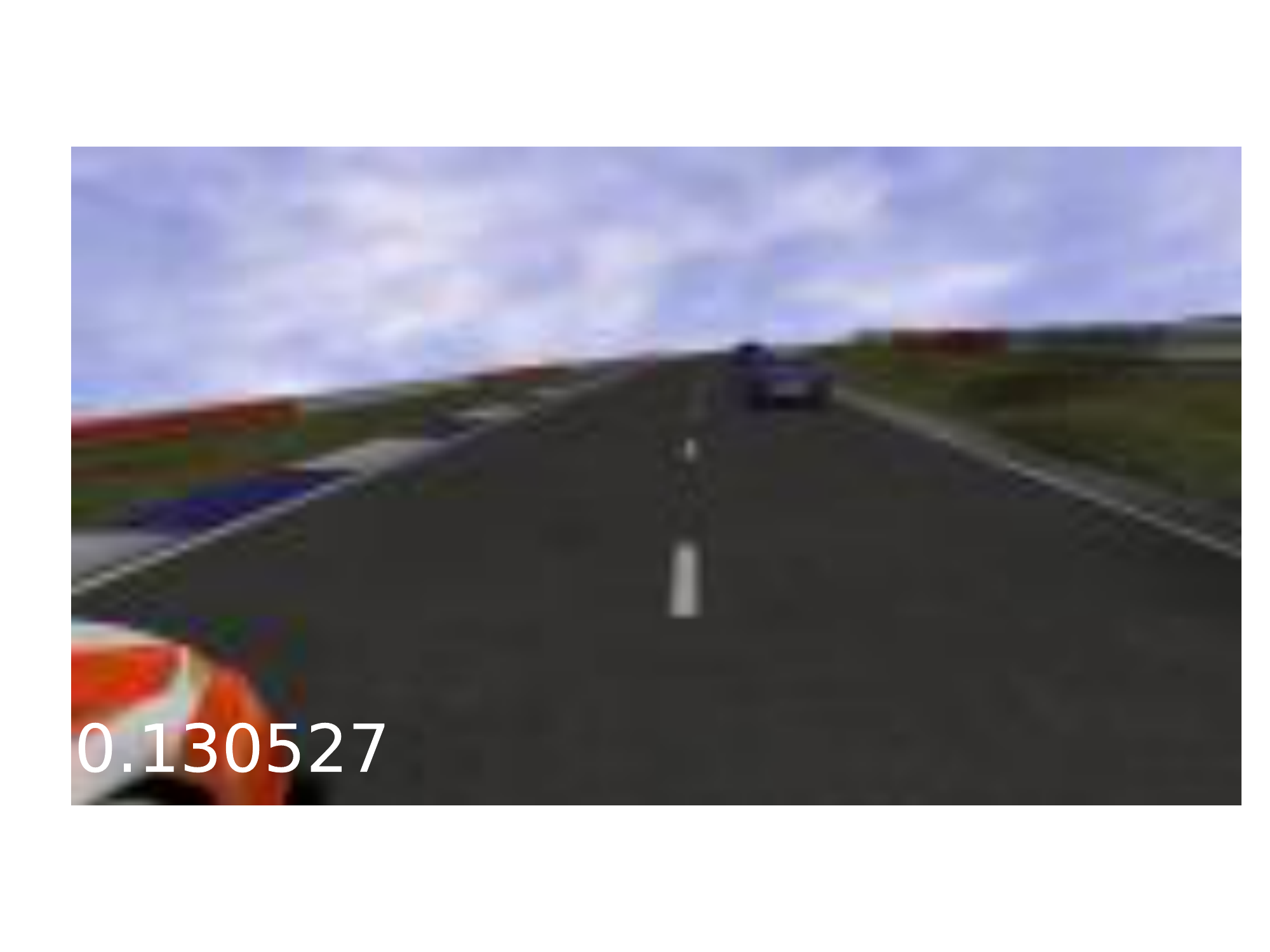}
        \unsafeframe{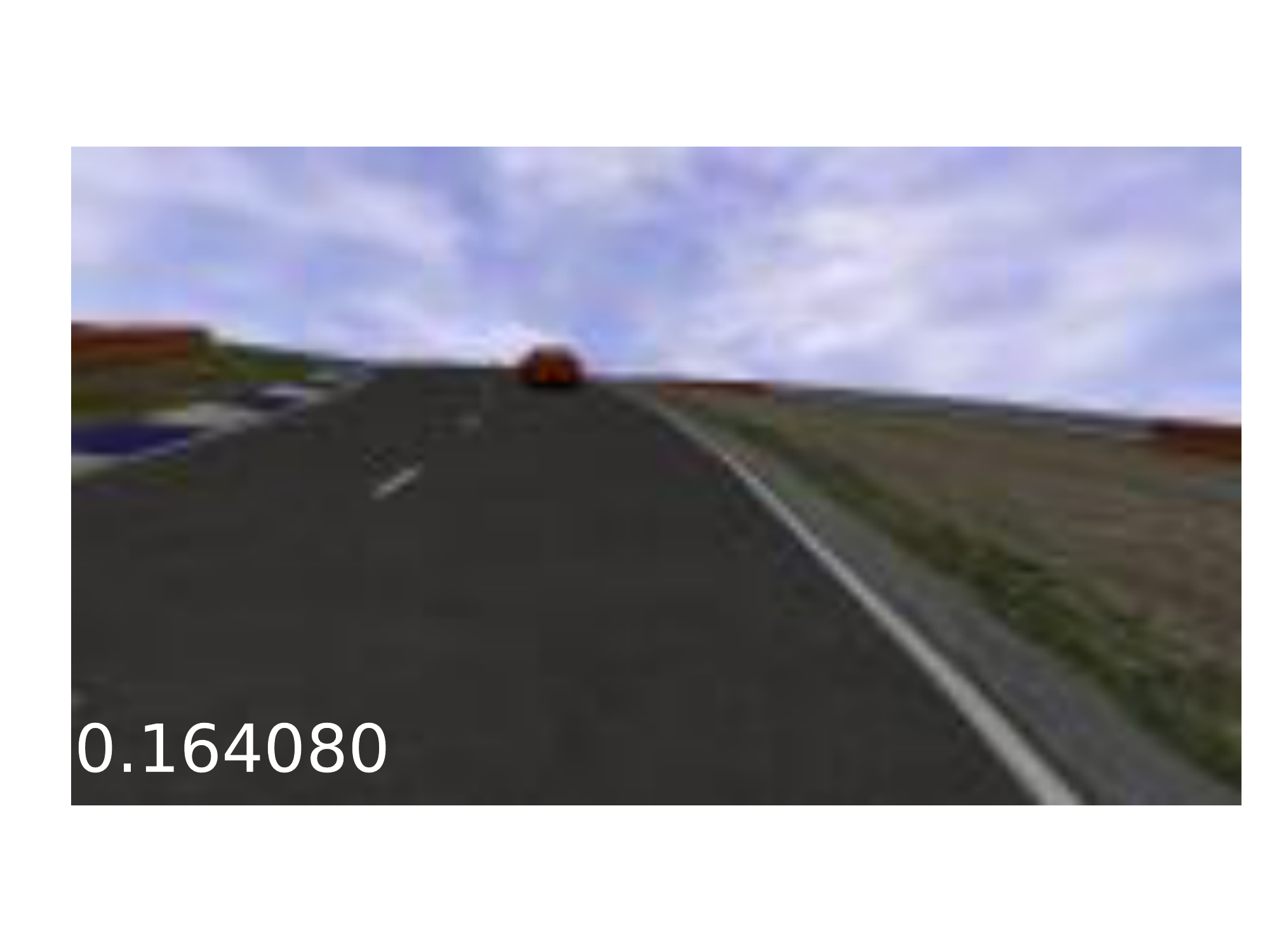}
    \end{minipage}

    \caption{Sample image frames sorted according to a safety policy trained on
    a primary policy right after supervised learning stage. The number in each
frame is the probability of the safety policy returning $1$.}
\label{fig:frames}
\end{figure}

In Fig.~\ref{fig:frames}, we present twenty sample frames. The top ten frames
were considered {\it safe} (0) by a trained safety policy, while the bottom ones
were considered {\it unsafe} (1). It seems that the safety policy at this point
determines the safety of a current state observation based on two criteria; (1)
the existence of other cars, and (2) entering a sharp curve.

\end{document}